\documentclass[final]{cvpr}

\usepackage{times}
\usepackage{epsfig}
\usepackage{graphicx}
\usepackage{amsmath}
\usepackage{amssymb}
\usepackage{wrapfig}
\usepackage{xcolor}
\usepackage{multirow}
\usepackage{subfigure}
\usepackage{booktabs}

\newcommand{\CapRedistillation}{IE-KD}
\newcommand{\LowRedistilling}{IE-KD}
\newcommand{\LowRedistillation}{IE-KD}

\newcommand{\ieno}{\textit{i}.\textit{e}.}
\newcommand{\etcno}{\textit{etc}} 

\usepackage[pagebackref=true,breaklinks=true,colorlinks,bookmarks=false]{hyperref}

\def\cvprPaperID{941} 
\def\confYear{CVPR 2021}

\begin{document}

\title{Revisiting Knowledge Distillation: An Inheritance and Exploration Framework}


\author{
Zhen Huang\textsuperscript{1, 2}\thanks{This work was done when the author was visiting Alibaba as a research intern.},
Xu Shen\textsuperscript{2},
Jun Xing\textsuperscript{3},
Tongliang Liu\textsuperscript{4},
Xinmei Tian\textsuperscript{1}\thanks{Corresponding author.},\\
Houqiang Li\textsuperscript{1},
Bing Deng\textsuperscript{2},
Jianqiang Huang\textsuperscript{2},
Xian-Sheng Hua\textsuperscript{2}\footnotemark[2]\\
\textsuperscript{1}University of Science and Technology of China, 
\textsuperscript{2}Alibaba Group\\
\textsuperscript{3}University of Southern California,
\textsuperscript{4}University of Sydney\\
{\tt\small hz13@mail.ustc.edu.cn, junxnui@gmail.com, tongliang.liu@sydney.edu.au, \{xinmei,lihq\}@ustc.edu.cn,}\\
{\tt\small \{shenxu.sx,dengbing.db,jianqiang.hjq,xiansheng.hxs\}@alibaba-inc.com}
}

\maketitle

\pagestyle{empty}  
\thispagestyle{empty} 

\begin{abstract}
    Knowledge Distillation (KD) is a popular technique to transfer knowledge from a teacher model or ensemble to a student model. Its success is generally attributed to the privileged information on similarities/consistency between the class distributions or intermediate feature representations of the teacher model and the student model. However, directly pushing the student model to mimic the probabilities/features of the teacher model to a large extent limits the student model in learning undiscovered knowledge/features. In this paper, we propose a novel inheritance and exploration knowledge distillation framework (IE-KD), in which a student model is split into two parts - inheritance and exploration. The inheritance part is learned with a similarity loss to transfer the existing learned knowledge from the teacher model to the student model, while the exploration part is encouraged to learn representations different from the inherited ones with a dis-similarity loss. Our IE-KD framework is generic and can be easily combined with existing distillation or mutual learning methods for training deep neural networks. Extensive experiments demonstrate that these two parts can jointly push the student model to learn more diversified and effective representations, and our IE-KD can be a general technique to improve the student network to achieve SOTA performance. Furthermore, by applying our IE-KD to the training of two networks, the performance of both can be improved w.r.t. deep mutual learning. The code and models of IE-KD will be make publicly available at \url{https://github.com/yellowtownhz/IE-KD}. 
\end{abstract}

\section{Introduction}
\label{sec:introduction}
Knowledge distillation is one of the most popular methods for transferring
knowledge from one network (teacher) to another (student).  It was first
proposed by Hinton~\etal~\cite{Hinton2015DistillingTK} to transfer knowledge
from a large teacher network (or ensemble) to a small student network that is
easier to deploy.  It works by training the student to predict the target
classification labels and mimic the class probabilities of the teacher, as these features contain
additional information about how the teacher tends to generalize
\cite{Hinton2015DistillingTK}. All recent distillation works follow this
philosophy of an additional consistency control between the class probabilities
or intermediate representations of the teacher network and the student network.
KD \cite{Hinton2015DistillingTK} and Tf-KD \cite{KD-TF} focus on the
consistency of output class probabilities. AT \cite{Zagoruyko2016PayingMA}, AB
\cite{AB}, FT \cite{Kim2018ParaphrasingCN}, OD \cite{OD}, FEED \cite{FEED} and
FitNet \cite{Romero2015FitNetsHF} propose different consistency controls of intermediate
features. FSP \cite{fastop} proposes a consistency control of the
intra-similarities among intermediate features. In summary, all recent distillation methods differ in the metric of consistency between the student model and the teacher model.

\begin{figure*}[t]
    \centering
    \subfigure[Visualization of learned knowledge]
    {
        \includegraphics[width=0.33\linewidth]{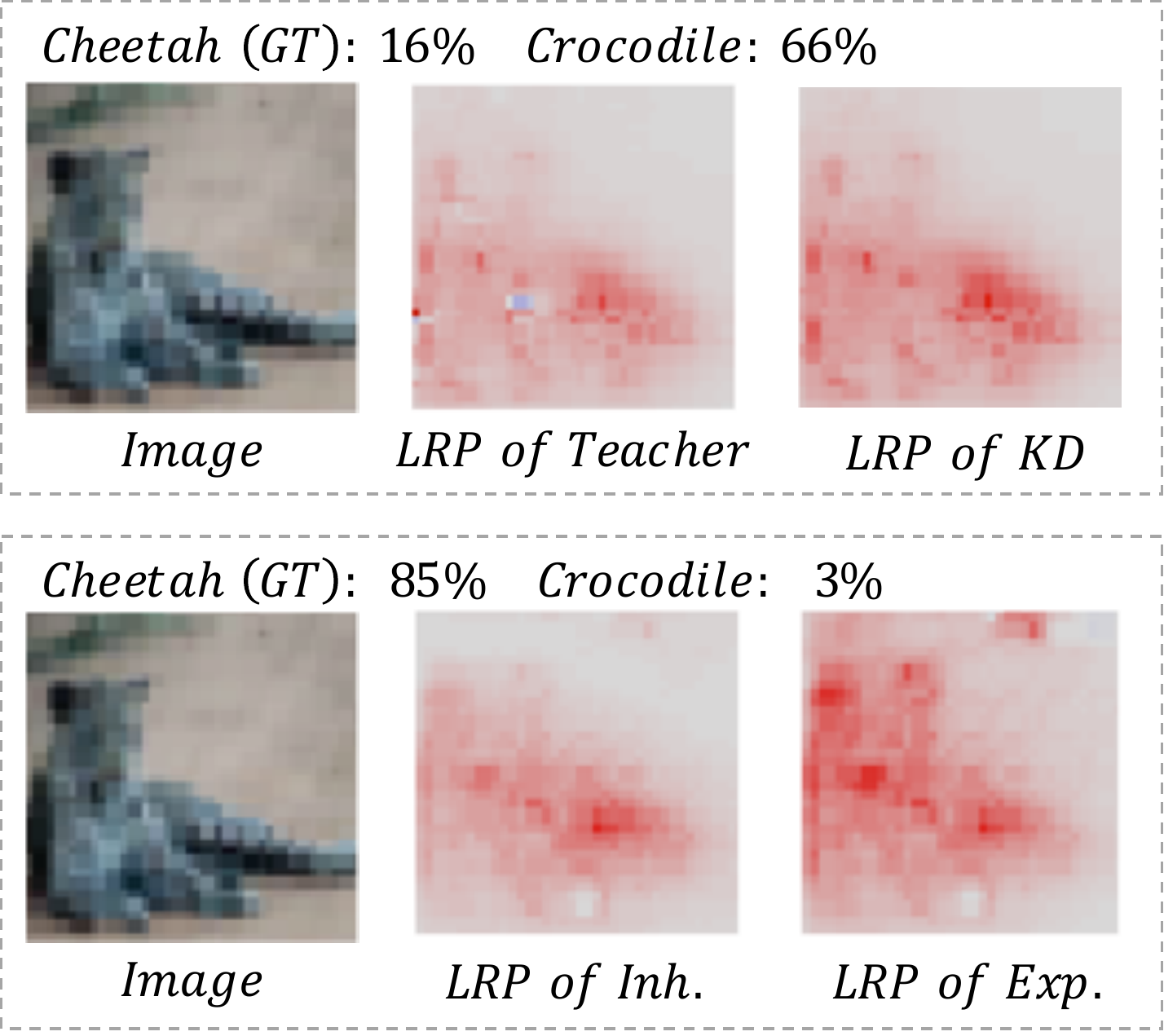}
        \label{fig:intro_LRP}
    }
    \subfigure[Loss curves]
    {
        \includegraphics[width=0.4\linewidth]{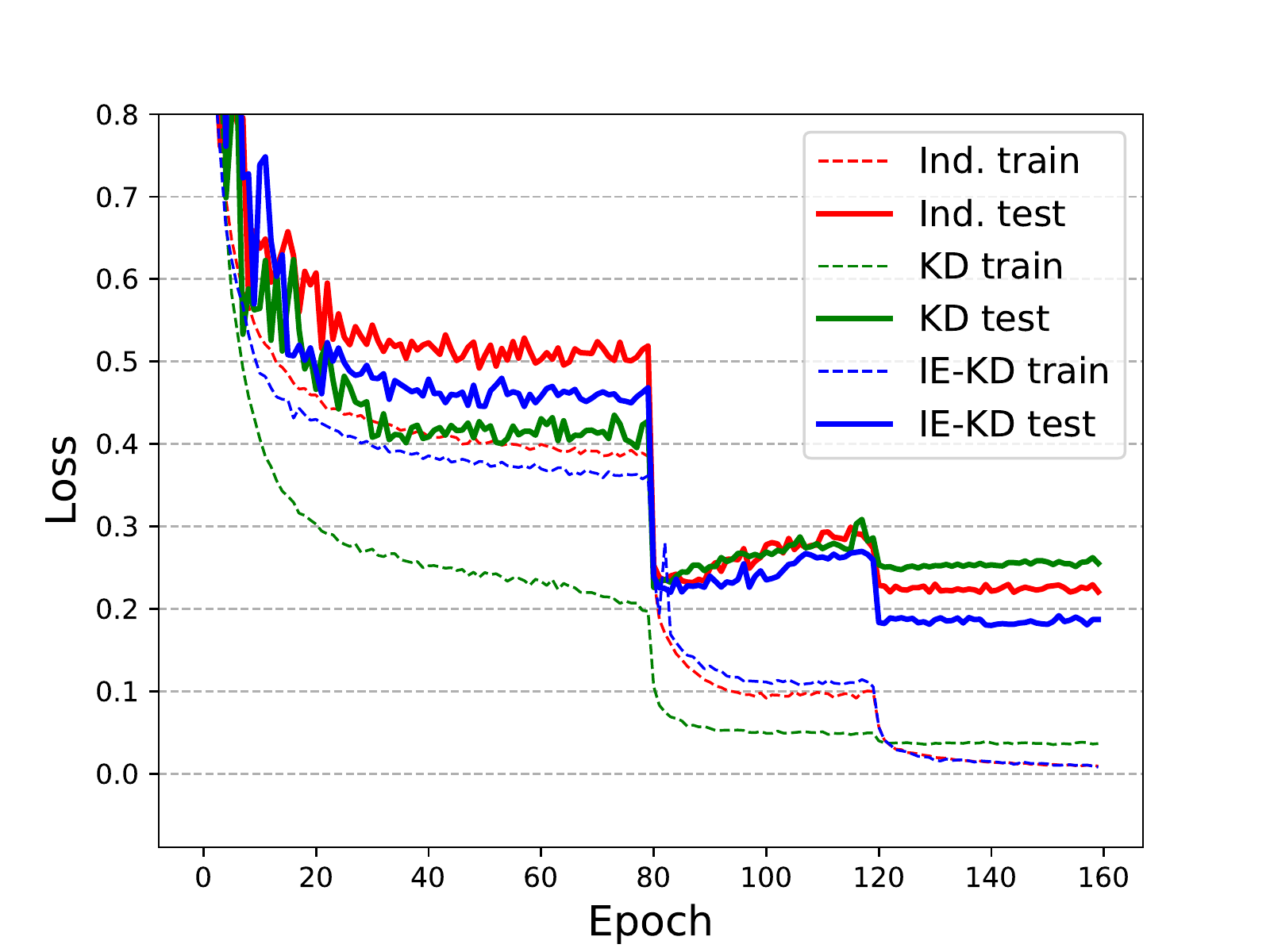}
        \label{fig:intro_loss}
    }
    \caption{
        \textbf{Left}: Visualization of learned knowledge for classification, including
        the teacher network (LRP of Teacher), student network trained with KD
        (LRP of KD), inheritance part of IE-KD
        (LRP of Inh.) and exploration part of IE-KD (LRP of Exp.).
        LRP \cite{lrpOverview} is used to interpret the network by visualizing which pixels contribute how much to the classification.
        \textbf{Right}: Training loss (dotted lines) and testing loss (bold lines) on CIFAR-10
        of the student network (ResNet-56) which is trained via independent learning
        (training-from-scratch), KD and IE-KD
        (using ResNet-20 as teacher network).
        For a fair comparison, KD and IE-KD correspond to FT and IE-FT here.
        Directly pushing the student network to mimic the outputs of the teacher network limits the student network in learning new knowledge. It even leads to a poor solution when the student network is larger than the teacher network (high training and testing loss at the same time).
    }
    \label{fig:intro}
\end{figure*}

However, directly pushing the student model to mimic the probabilities/features
of the teacher model limits the student model in learning new
knowledge/features. As shown in Fig. \ref{fig:intro_LRP},
the student model trained with KD learns very similar patterns compared
with the well-trained teacher (more results will be shown in supplementary
materials).
In this case, the ``cheetah'' misclassified as a ``crocodile'' by the
teacher model is also misclassified by the student model trained by KD. The model attributes most
of its prediction to the tail of the ``cheetah'' which resembles a ``crocodile''.
As a result, the student network fails to incorporate new relevant patterns on ears and mouth that are quite discriminative between the ``cheetah'' and ``crocodile''.
Therefore, we need a mechanism to find more useful features for correct predictions that are omitted by the teacher network.

Intuitively, simply mimicking outputs of the teacher network  will narrow the
search space for the optimal parameters of the student network and lead to
a
poor solution from a feature learning view.
Furthermore, we find that this phenomenon becomes more evident when transferring knowledge from a small teacher network to a large student network (shown in Fig. \ref{fig:intro_loss}).
According to the observation in \cite{Ba2013DoDN,ModelCompression}, small networks often have as sufficient capacity as large networks but represent the features in a more concise manner \cite{Romero2015FitNetsHF}.  Therefore, a large network should not only mimic this compact representation with some of their parameters to reduce the redundancy of itself, but also should free other parameters to explore more different and complementary features to improve its diversity and generalization ability.
Based on the aforementioned analyses, in this paper, we propose a novel inheritance and exploration knowledge distillation framework (\textit{IE-KD}), to train a student network by partially following the knowledge from the teacher network and partially exploring for new knowledge that are complementary to the teacher network.

In our IE-KD, the knowledge is transferred by the two principles of consistency
and diversity.  Consistency ensures that the well learned knowledge encoded in
the teacher network is successfully inherited by the student network.
Diversity ensures that the student network can explore new features that are
complementary to the inherited ones.  The motivation of IE-KD comes from the
theory of heredity in evolution \cite{darwin1859}. Heredity involves
inheritance and variation of traits.  Evolution results from natural selection
acting on diversity in populations, which originally stems from mutations.
There are three key factors for evolution: a) inheritance of compact and
effective traits from parents encoded by genes, b) new diversified genotypes
generated from genetic mutations, and c) natural selection through stressful
environments.  Motivated by this, we split the student network into two parts:
one inherits the compact and effective knowledge encoded by factors from the
teacher network via consistency/inheritance loss (similarity), and the other is
pushed to generate different features via diversity/exploration loss
(dis-similarity).  The supervised task (classification/detection) loss plays
the role of natural selection, guiding the exploration part to converge to
diverse yet effective features.

Another closely related motivation for
\LowRedistillation{} comes from the exploration of actions in Q-learning
\cite{MnihKSGAWR13}, and the popular AlphaGo \cite{alphago}, where half the
actions follow the predictions of the policy network, and the other half are
randomly sampled from the remaining action space that ensures adequate
exploration of the state space.
Besides, \cite{chen2020universal} proposes a similar form of loss function to attack the heat maps of one white-box DNN, making its attention focus on other regions of the image.
Inspired by these insights, we propose our IE-KD framework to improve the training of student network, by exploring the new and undiscovered knowledge apart from the teacher-learned knowledge.

Overall, our IE-KD framework is generic and can be easily combined with
existing distillation or mutual learning methods for training deep neural
networks. Extensive experiments demonstrate that these two parts can jointly
push the student model to learn more diversified and effective representations,
and our \CapRedistillation{} can be a general technique to improve the student
network to achieve SOTA performance. Furthermore, by applying our IE-KD to the
training of two networks, the performance of both can be improved w.r.t. deep mutual learning.

\section{Related Work}
\label{sec:relatedWorks}
In this paper, we focus on knowledge transfer between networks.  All related
works can be divided into three types: consistency control from a pre-trained
teacher network to a student network by distillation, simultaneous learning of
network pairs by consistency control, and self-distillation by teacher free
regularization.

\textbf{Consistency Control from a pre-trained teacher network or ensemble to
a student network}.
Various approaches exist to transfer knowledge from a pre-trained large network
or ensemble to an untrained small network, \ieno, knowledge distillation.  The
transferred knowledge lies in a consistency of output probabilities
(KD \cite{Hinton2015DistillingTK}), intermediate features
(AT \cite{Zagoruyko2016PayingMA}, AB \cite{AB}, FT \cite{Kim2018ParaphrasingCN},
OD \cite{OD}, FEED \cite{FEED}, FitNet \cite{Romero2015FitNetsHF}),
or similarities between intermediate features (FSP \cite{fastop}). Each method
differs in the metric of consistency, including KL divergence between output
probabilities (KD \cite{Hinton2015DistillingTK}, BAN \cite{BAN}), regression
with additional parameters between the mapping of intermediate features (FitNet
\cite{Romero2015FitNetsHF}), $L1$ distance between projected factors (FT
\cite{Kim2018ParaphrasingCN}), $L1$ distance between the pooled attentions (AB
\cite{AB}), and $L2$ distance between rectified activations (OD \cite{OD}).
FEED \cite{FEED} proposed $L1$ distances between the features of an ensemble of
teacher networks and the untrained small network.
CRD \cite{CRD} proposed a contrastive-based objective for transferring high-order dependencies in representational space between deep networks.

\textbf{Simultaneous learning by consistency control among a group of untrained
networks}.
Recently, researchers have proposed to relax the requirements of a
pre-trained large network by starting with a pool of untrained networks and
learns simultaneously with a consistency control. Deep mutual learning
\cite{Zhang2017DeepML} shows that an ensemble of students could learn
collaboratively and teach each other throughout the training process by
consistency control of output probabilities. More recently, FFT
\cite{Kim2019FeatureFF}, ONE \cite{KD-on-the-fly} and CL \cite{CL} proposed
consistency control between an ensemble of sub-network classifiers and each
sub-network, where each sub-network mutually teaches one another in an
online-knowledge distillation manner.

\textbf{Teacher-free regularization}.
In Tf-KD \cite{KD-TF}, label smoothing regularization was introduced as a
virtual teacher model for KD, without any additional peer networks needed. SD
\cite{SnapshotKD} proposed to use snapshots from earlier epochs as teacher
model. These works still comply to the consistency between student network and
referred targets, either manually designed or selected from snapshots.

In this study, we propose a new framework for transferring knowledge from the
teacher network to a student network. Beyond the consistency control used in
distillation and mutual learning, \LowRedistillation{} further involves a
diversity control.  In addition, our \LowRedistilling{} approach supports
similar mutual learning between a group of networks and achieves much better
performance.


\section{Method}
\label{sec:method}

\begin{figure*}[t]
    \vspace{-1mm}
    \centering
    \includegraphics[width=0.9\linewidth]{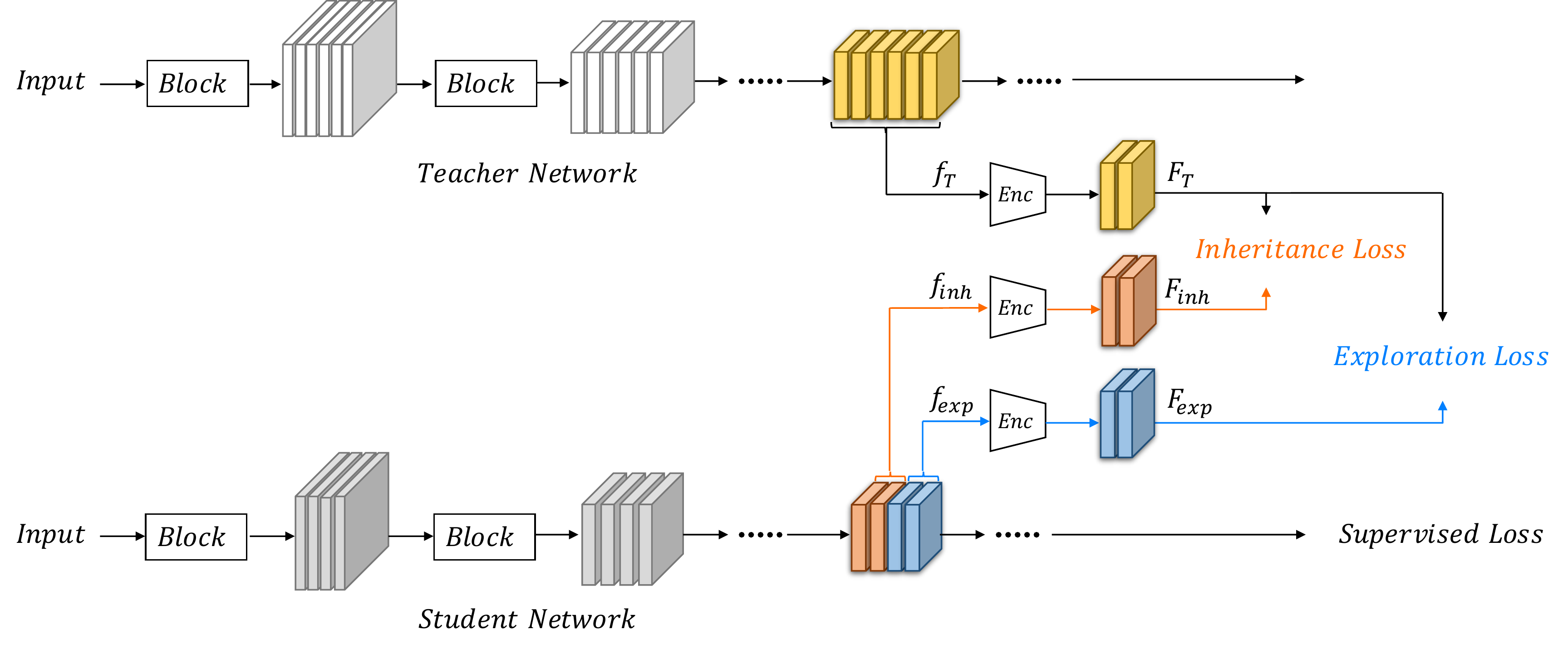}
    \caption{
        Overview of \LowRedistilling{} framework.
        The student network is split into two parts.
        One part (colored in \textcolor{orange}{orange}) inherits the compact
        and effective representations encoded by factors from the teacher network via
        consistency/inheritance loss (similarity),
        and the other part (colored in \textcolor{blue}{blue}) is pushed to
        generate different features via diversity/exploration loss
        (dis-similarity).
        The supervised task (classification/detection) loss guides the
        exploration part to converge to diverse yet effective features.
    }
    \label{fig:framework}
    \vspace{-3mm}
\end{figure*}

Fig. \ref{fig:framework} illustrates the framework of our approach.  The features of
the student network is divided into two parts. One part (indicated by
the orange color) is trained to mimic the compact features of the teacher
network using an inheritance loss, and the other part (blue) is
encouraged to learn new features different from the teacher network via an
exploration loss.
The supervised task (classification/detection) loss guides the exploration part to converge to diverse yet effective features.
Overall, the student network is
trained with the inheritance loss and the exploration loss, together with the
conventional supervised target loss.

Since the teacher network is pre-trained, the compact features could be
pre-learned as well using auto-encoder, which we will discuss in
Sec. \ref{sec:student_extraction}. Then, we will discuss the details of
\LowRedistilling{} in Sec. \ref{sec:dedistill}, followed by extension to deep mutual learning manner in Sec. \ref{sec:mutual_learning}.

\subsection{Compact Knowledge Extraction}
\label{sec:student_extraction}

We denote the features of the teacher as $f_T$, the features of the inheritance part  and exploration part of the student network as  $f_{inh}$ and  $F_{exp}$, respectively. The challenge in measuring the similarity/dis-similarity between these features is that they usually have different shapes and sizes. To solve this problem, we embed
them into a shared latent feature space of the same dimension via an encoder, and the embedded features are indicated by
$F_T$, $F_{inh}$ and $F_{exp}$, respectively. 
We adopt the factor-based embedding module in \cite{Kim2018ParaphrasingCN} to
extract knowledge from the specific convolutional block of the teacher network.

In particular, an auto-encoder, consisting of several convolutional and
deconvolutional layers, is adopted to extract transferable factors $F_T$ from
the teacher network.
We use three convolution layers
and three transposed convolution layers. All six layers use $3\times3$
kernels, stride of $1$, padding of $1$, and batch normalization with
leaky-ReLU with rate of $0.1$ followed by each of the six layers.
Only at the
second convolution, the number of output feature maps is compressed to the number of 
factor feature maps. Similarly, the second transposed convolution layer is
resized to match the feature maps of the teacher network. The detailed architecture can be found in the supplementary materials.
The auto-encoder is trained by the common reconstruction loss:
\vspace{-2mm}
\begin{equation}
    L_{rec} = ||f_{T} - R(f_{T})||^2,
    \label{eqn:L_rec}
\vspace{-1mm}
\end{equation}
where $f_{T}$ is the feature maps of the teacher
network and $R(f_{T})$ is the output of the auto-encoder.

\subsection{Inheritance and Exploration}
\label{sec:dedistill}

The goal of \LowRedistillation{} is to enhance the features of the student
network, $f_{S}$, by using the compact features of the teacher network, $f_{T}$.
Directly pushing the student model to mimic the features of the teacher model limits the student model in learning undiscovered features.
Therefore, instead of treating and training $f_{S}$ as a whole, we randomly split it into
two parts, $f_{inh}$ and $f_{exp}$, and regulate them separately with two
counterpart losses, an inheritance loss ${L}_{inh}$ that pushes $f_{inh}$ to
mimic $f_{T}$ as much as possible, and an exploration loss ${L}_{exp}$ that
allows $f_{exp}$ to learn different or unrelated features to $f_{T}$.
Similarly, we use two separate encoders to embed $f_{inh}$ and $f_{exp}$ into
factors $F_{inh}$ and $F_{exp}$ that have the same dimension as $F_{T}$, which
relieves the student network from the burden of directly learning the output of
the teacher network.

We would like to emphasize that the specific implementation of the inheritance
loss and exploration loss are orthogonal to our \LowRedistillation{} framework.
\emph{All metrics that measures the similarity and dis-similarity of two representations
can be easily adopted into our framework for the inheritance part and exploration
part, respectively.}
Below, we only discuss a simple and effective implementation of ${L}_{inh}$ and
${L}_{exp}$.

{\bf Inheritance loss.} ${L}_{inh}$ is designed to inherit the existing knowledge from the teacher model by minimizing the difference between $F_{inh}$ and $F_{T}$, and is represented as:
\vspace{-2mm}
\begin{equation}
    \mathcal{L}_{inh} = ||\frac{F_{inh}}{||F_{inh}||_2} -
    \frac{F_{T}}{||F_{T}||_2}||_1.
    \label{eqn:L_inh}
    \vspace{-1mm}
\end{equation}
Similar to \cite{Kim2018ParaphrasingCN}, we apply $L_1$ normalization to the
factors.  The $L_1$ distance acts as a similarity metric in a very simple form.
Any other similarity metrics for vectors can be easily adopted as a inheritance
loss, such as $L_2$, cosine distance ($L=1-cos(x, y)$), partial $L_2$ distance
\cite{OD}, \etcno.

{\bf Exploration loss.} ${L}_{exp}$ is designed to act oppositely as
${L}_{inh}$, learning representations that are different from the inherited ones. Inspired by \cite{chen2020universal}, a straightforward choice is to minimize the negative
difference between $F_{exp}$ and $F_{T}$:
\vspace{-2mm}
\begin{equation}
    \mathcal{L}_{exp} = - ||\frac{F_{exp}}{||F_{exp}||_2} -
    \frac{F_{T}}{||F_{T}||_2}||_1.
    \label{eqn:L_exp}
    \vspace{-1mm}
\end{equation}
We would like to point out that the sign change of the exploration loss is
different from pushing $F_{exp}$ to learn a negative teacher factor $-F_{T}$,
which obviously correlates with $F_{T}$.
\emph{${L}_{exp}$ aims to encourage the exploration part to focus on other regions of the image \cite{chen2020universal}, exploring new features that are complementary to the inherited ones.}

Likewise, there exist many different
metrics to measure the dis-similarity, such as negative $L_2$ distance
($L=-||x-y||_2$), orthogonality measure ($L=|cos(x, y)|$), CKA \cite{cka},
negative partial $L_2$ distance \cite{OD}, \etcno.

{\bf Training.} The teacher network's factor auto-encoder is firstly trained
with the reconstruction loss. Then, the factor encoders and backbone network of
the student network is trained simultaneously with target loss (classification,
detection, \etcno.), inheritance loss and exploration loss:
\vspace{-2mm}
\begin{equation}
    \mathcal{L} = \mathcal{L}_{goal} + \lambda_{inh}
    \mathcal{L}_{inh} + \lambda_{exp} \mathcal{L}_{exp},
    \vspace{-2mm}
\end{equation}
where $\lambda_{inh}$ and $\lambda_{exp}$ are the corresponding loss weights, respectively.

\subsection{Extension to Deep Mutual Learning}
\label{sec:mutual_learning}

In the above sections, we propose a new framework to improve a student network by transferring knowledge from a teacher network in an inheritance and exploration manner.
A straightforward idea is that we can further improve the teacher network via the same process with the better student network.
Thus, our IE-KD approach can be extended to a deep mutual learning manner \cite{Zhang2017DeepML} (termed as \textit{IE-DML}), to make both the teacher and student networks benefit from our IE-KD mechanism.

In the original Deep Mutual Learning \cite{Zhang2017DeepML} strategy, two peer networks ($\Theta_1$ and $\Theta_2$) are optimized simultaneously with KL distance to measure the consistency of two network's predictions.
In IE-DML, we replace the KL regularization with two IE-KD processes (described in Sec. \ref{sec:dedistill}) which have opposite directions, \ieno, network $\Theta_1$ to $\Theta_2$ and network $\Theta_2$ to $\Theta_1$.
For example, in the process of network $\Theta_1$ to network $\Theta_2$, network $\Theta_2$ is trained by regarding network $\Theta_1$ as the teacher network.
Additionally, since both networks are trained from scratch, the auto-encoder needs to be trained together with the backbone networks.

The overall loss function $\mathcal{L}_{\Theta_1}$ for network $\Theta_1$ is composed of four components: target loss, reconstruction loss, inheritance loss and exploration loss:
\vspace{-2mm}
\begin{equation}
    \mathcal{L}_{\Theta_1} = \mathcal{L}_{goal1} + \lambda_{rec} \mathcal{L}_{rec1} + \lambda_{inh} \mathcal{L}_{inh1} + \lambda_{exp} \mathcal{L}_{exp1},
\label{eqn:DML1}
\vspace{-2mm}
\end{equation}
where the $\lambda_{rec}$, $\lambda_{inh}$ and $\lambda_{exp}$ are the corresponding loss weights, respectively.
Similarly, the objective loss function $\mathcal{L}_{\Theta_2}$ for network $\Theta_2$ can be computed as:
\vspace{-2mm}
\begin{equation}
    \mathcal{L}_{\Theta_2} = \mathcal{L}_{goal2} + \lambda_{rec} \mathcal{L}_{rec2} + \lambda_{inh} \mathcal{L}_{inh2} + \lambda_{exp} \mathcal{L}_{exp2}.
\label{eqn:DML2}
\vspace{-2mm}
\end{equation}
In this way, each model is trained by the inheritance and exploration with the compact knowledge from the other one.

Finally, two networks are updated alternatively following four steps until convergence:
a) update the predictions of the teacher and student networks for an input mini-batch;
b) compute the stochastic gradient w.r.t. $\mathcal{L}_{\Theta_1}$, and update $\Theta_1$;
c) update the predictions of the teacher and student networks for the current mini-batch;
d) compute the stochastic gradient w.r.t. $\mathcal{L}_{\Theta_2}$, and update $\Theta_2$.

\section{Experiments}
\label{sec:experiments}

The efficiency of our IE-KD mechanism is evaluated on both classification and
detection tasks.  For classification, CIFAR and ImageNet datasets are used. For
detection, the PASCAL VOC dataset is used.  We compare our proposed IE-KD with
independent learning and several state-of-the-art knowledge distillation
methods.  In independent learning, both the teacher and student networks are
independently trained from scratch.  In distillation, the student network is
trained by transferring knowledge from the pre-trained teacher network via
consistency controls.  By comparing with distillation, we demonstrate that our
\LowRedistillation{} is a general method to improve the student network to
achieve SOTA performance and our exploration loss plays a key role in enhancing the network features.

\subsection{Datasets and Settings}
\label{sec:settings}

CIFAR-10 and CIFAR-100 \cite{krizhevsky2010cifar} consist of $50,000$ training
and $10,000$ test images drawn from $10$ and $100$ classes.
Networks are trained using SGD with Nesterov momentum. The initial learning
rate is set to $0.1$, the momentum is set to $0.9$, and the mini-batch size is
set to $128$. The learning rate is divided by $10$ at the $80$th and $120$th
epochs. The training process ends at the $160$th epoch.

ImageNet consists of $1.2$M training images and $50$k validation images with
$1,000$ classes.  We perform large-scale experiments on ImageNet
to verify our potential ability to transfer more complex and detailed
information. Networks are trained for $100$ epochs. The learning rate
begins at $0.1$ and is multiplied by $0.1$ at every $30$ epochs.

We apply our method to Single Shot Detector (SSD) \cite{liu2016ssd}.  Networks
are trained on a mixture of the PASCAL VOC$2007$ and VOC$2012$
\cite{everingham2015pascal} \textit{trainval} sets, which are widely used in
object detection. The backbone network in all models is pre-trained using
ImageNet. Networks are trained for $120$k iterations with a batch size of $32$.
Detection performance is evaluated on the VOC $2007$ \textit{test} set.

\subsection{Implementation Details}
We implement all networks in PyTorch \cite{pytorch} and the code
will be released later. The ratio of the number of input feature maps to the
number of factor feature maps is set to $2$. We randomly split representations
of the student network into inheritance and exploration parts, since the
parameters of the network are randomly initialized and there is no strong
correlation among channels before learning. The weights of both inheritance and
exploration loss are set to $50$ on CIFAR-10, $100$ on ImageNet and PASCAL VOC.

Our IE-KD approach is a general framework and can be easily combined with
existing distillation methods.  In this paper, we combine our IE-KD framework with 
three SOTA distillation methods, AT
\cite{Zagoruyko2016PayingMA}, FT \cite{Kim2018ParaphrasingCN} and OD \cite{OD} , and denote them as
IE-AT, IE-FT and IE-OD, respectively.
In Sec. \ref{sec:dedistill}, we present the formulation of IE-FT as an
instantiation of our approach.
For IE-AT, the output factors of the encoders are reduced to spatial
attention maps first, then the inheritance loss (Eq.(\ref{eqn:L_inh})) and exploration loss (Eq.(\ref{eqn:L_exp})) are applied to the
spatial attention maps between the teacher network and the
inheritance/exploration part of the student network.
For IE-OD, we revise the distance formulations as in OD \cite{OD}, \ieno, Eq.(\ref{eqn:L_inh}) is reformulated as $\mathcal{L}_{inh}=||(max(F_{inh}, 0) - max(F_S, 0)||_2$ and Eq.(\ref{eqn:L_exp}) becomes $\mathcal{L}_{exp}=-||(max(F_{exp}, 0) - max(F_S, 0)||_2$.

\subsection{Results of Image Classification}
\label{sec:result_classification}
\begin{table*}[htpb]
\vspace{-1mm}
\centering
\caption{
    KD \cite{Hinton2015DistillingTK}, AT \cite{Zagoruyko2016PayingMA}, FT
    \cite{Kim2018ParaphrasingCN}, OD \cite{OD}, Tf-KD \cite{KD-TF}, CRD
    \cite{CRD} and IE-KD experiments results by training the student
    network with pre-trained teacher (error, in \%) on CIFAR-10.
}
\begin{tabular}{cc|cccccc|ccc}
    \toprule
    Teacher: baseline & Student: baseline & KD  &  AT  &  FT  &  OD  & Tf-KD (S) & CRD  & IE-AT & IE-FT & IE-OD \\
    \midrule
    ResNet-56: 6.39 & ResNet-20: 7.78 & 7.37 & 7.13 & 6.85 & 6.81 & 7.41 &
    6.80 & 6.70 & 6.57 & \textbf{6.53} \\
    WRN-40-1: 6.84 & ResNet-20: 7.78 & 7.46 & 7.14 & 6.85 & 6.69 & 7.51 &
    6.77  & 6.81 & 6.57 & \textbf{6.49} \\
    WRN-46-4: 4.44 & VGG-13:\quad 5.99 & 5.59 & 5.48 & 4.84 & 4.81 & 5.48 & 4.81 &
    4.75 & 4.67 & \textbf{4.65} \\
    WRN-16-2: 6.27 & WRN-16-1: 8.62 & 8.22 & 8.01 & 7.64 & 7.48 & 8.10 & 7.49
    & 7.76 & 7.38 & \textbf{7.26} \\
    \bottomrule
\end{tabular}
    \label{tab:cifar10-student}
    \vspace{-1mm}
\end{table*}

\begin{table*}[htpb]
\centering
\caption{
    KD \cite{Hinton2015DistillingTK}, AT \cite{Zagoruyko2016PayingMA}, FT
    \cite{Kim2018ParaphrasingCN}, OD \cite{OD}, Tf-KD \cite{KD-TF}, CRD
    \cite{CRD} and IE-KD experiments results by training the student
    network with pre-trained teacher (Top-$1$ and Top-$5$ error, in \%) on
    ImageNet. The teacher is ResNet-34 and the student is ResNet-18.
}
\begin{tabular}{c|cc|cccccc|ccc}
    \toprule
    & Teacher & Student & KD & AT & FT & OD & Tf-KD (S) & CRD & IE-AT & IE-FT & IE-OD \\
    \midrule
    Top-1 & 26.73 & 29.91 & 29.34 & 29.30 & 28.57 & 28.49 & 29.58 & 28.83 &
    28.41 & 28.27 & \textbf{28.19} \\
    Top-5 & 8.57 & 10.68 & 10.12 & 10.00 & 9.71 & 9.67 & 10.06 & 9.87 & 9.54 &
    9.39 & \textbf{9.33} \\
    \bottomrule
\end{tabular}
\label{tab:imagenet-student}
\vspace{-1mm}
\end{table*}

\begin{table*}[!h]
\centering
\caption{
    AT \cite{Zagoruyko2016PayingMA}, FT \cite{Kim2018ParaphrasingCN}, OD
    \cite{OD}, CRD \cite{CRD} and IE-KD experiments results by training the
    student network with pre-trained teacher (mAP, in \%) on PASCAL VOC2007.
    The teacher is ResNet-50 and the student is ResNet-18.
}
\begin{tabular}{cc|cccc|ccc}
    \toprule
    Teacher & Student & AT & FT & OD & CRD & IE-AT & IE-FT & IE-OD \\
    \midrule
    76.79 & 71.61 & 72.00 & 72.68 & 73.08 & 73.11 & 73.16 & 73.32 & \textbf{73.51} \\
    \bottomrule
\end{tabular}
\label{tab:detection-student}
\vspace{-3mm}
\end{table*}

\begin{table*}[htpb]
    \vspace{-1mm}
    \centering
    \caption{
        Comparison of top-$1$ error (\%) on CIFAR-100 between DML
        \cite{Zhang2017DeepML} and our IE-DML.
    }
    \begin{tabular}{cc|cc|cc|cc}
        \toprule
        \multicolumn{2}{c|}{Network Types} &  \multicolumn{2}{c|}{Independent} &
        \multicolumn{2}{c|}{DML} & \multicolumn{2}{c}{IE-DML} \\
        Net 1 & Net 2  & Net 1 & Net 2 & Net 1 & Net 2 & Net 1 & Net 2 \\
        \midrule
        ResNet-32 & ResNet-110 & 31.01 & 26.91 & 28.69 & 25.59 & \textbf{28.36} & \textbf{24.99} \\
        ResNet-32 & WRN-28-10  & 31.01 & 21.31 & 29.27 & 21.04 & \textbf{28.06} & \textbf{20.63} \\
        ResNet-32 & MobileNet  & 31.01 & 26.35 & 28.90 & 23.87 & \textbf{28.33} & \textbf{23.24} \\
        ResNet-32 & ResNet-32  & 31.01 & 31.01 & 29.25 & 28.81 & \textbf{28.59} & \textbf{28.15} \\
        \bottomrule
    \end{tabular}
    \label{tab:cifar100-dml}
    \vspace{-2mm}
\end{table*}

\begin{table*}[!h]
    \centering
    \caption{
        Ablation study with different proportions of inheritance and exploration feature channels (top-$1$ error in \%).
    }
    \begin{tabular}{c|cc|ccccc}
        \toprule
        Dataset & Teacher & Student & $\eta=0.0/1.0$ & $0.3/0.7$ & $0.5/0.5$ & $0.7/0.3$ & $1.0/0.0$ \\
        \midrule
        CIFAR-10 & ResNet-56 & ResNet-20 & 7.56 & 6.84 & \textbf{6.53} & 6.71 & 6.86 \\
        CIFAR-100 & ResNet-110 & ResNet-32 & 28.01 & 27.23 & \textbf{25.67} & 26.37 & 27.49 \\
        ImageNet  & ResNet-34 & ResNet-18 & 30.02 & 28.90 & \textbf{28.19} & 28.64 & 29.65 \\
        \bottomrule
    \end{tabular}
    \label{tab:ablation_prop}
    \vspace{-3mm}
\end{table*}

To control other factors and make a fair comparison, we
reproduced the algorithms of other methods based on their codes and papers.
Table \ref{tab:cifar10-student} shows the Top-$1$ error rate on CIFAR-10 when
various architectures, including ResNet \cite{he2016deep}, Wide ResNet
\cite{zagoruyko2016wide} and VGG \cite{simonyan2014very},  are used.
In the table, the ``teacher: baseline'' and ``student: baseline'' columns
denote the network architecture and corresponding performance of training from
scratch.
First, we use ResNet-56 as teacher and ResNet-20 as student, that have same
number of channels but different blocks.
Then, we test different types of residual networks for teacher and student with
WRN-40-1 and ResNet-20.
To investigate the effect of the absence of shortcut connections, we further
use WRN-46-4 as teacher and VGG13 as student.
To test the applicability for architectures with the same blocks but different
channels, we use WRN-16-2 as teacher and WRN-16-1 as student.

Results of $3$ variants of IE-KD (IE-AT, IE-FT and IE-OD) and
other distillation methods on CIFAR-10 are presented in Table
\ref{tab:cifar10-student}.
We have two observations:
1) In all three variants of IE-KD, our inheritance and exploration framework
consistently outperforms corresponding consistency-based distillation method
(IE-AT vs. AT, IE-FT vs. FT, IE-OD vs. OD) with a significant margin. The
inheritance and exploration part can jointly push the student model to learn
more effective representations, resulting in a better performance.
2) all variants of IE-KD shows better performances than other latest
distillation methods (KD
\cite{Hinton2015DistillingTK}, AT \cite{Zagoruyko2016PayingMA}, FT
\cite{Kim2018ParaphrasingCN}, OD \cite{OD}, Tf-KD \cite{KD-TF} and CRD
\cite{CRD}) consistently, regardless of the type of teacher/student networks.
Furthermore, when faced with knowledge transfer from small teacher network to
large student network, our IE-KD shows even more improvement than other
distillation methods, results are presented in supplementary materials.


For further validation of generalization ability for large-scale image
classification task, we compare our IE-KD and other distillation methods on
ImageNet \cite{russakovsky2015imagenet}. Results are shown in
Table \ref{tab:imagenet-student}.  Following \cite{Kim2018ParaphrasingCN}, we
set ResNet-34 as a pre-trained teacher network and ResNet-18 as an untrained
student network. IE-KD outperforms all other methods again.

These results confirm that our IE-KD is a very general and effective
upgrade of existing distillation framework.

\subsection{Results of Object Detection}
\label{sec:result_detection}

We further verify the effectiveness of IE-KD for detection tasks. We
set ResNet-50 as the teacher network and ResNet-18 as the student network. Both
networks are pre-trained with ImageNet and fine-tuned on PASCAL VOC $2007$. As
shown in Table \ref{tab:detection-student}, with IE-KD from the teacher
network, the mean average precision (mAP) of the student network is increased
with a large margin ($71.61\%$ to $73.51\%$).
In this scenario, our IE-KD still shows a notable
improvement for the student network, showing that our method can be
applied to general computer vision tasks.

\subsection{Extension to Deep Mutual Learning}
\label{sec:exp_mutualTrain}



\begin{figure*}[htbp]
\vspace{-3mm}
    \centering
    \subfigure
    {
        \includegraphics[width=0.28\linewidth]{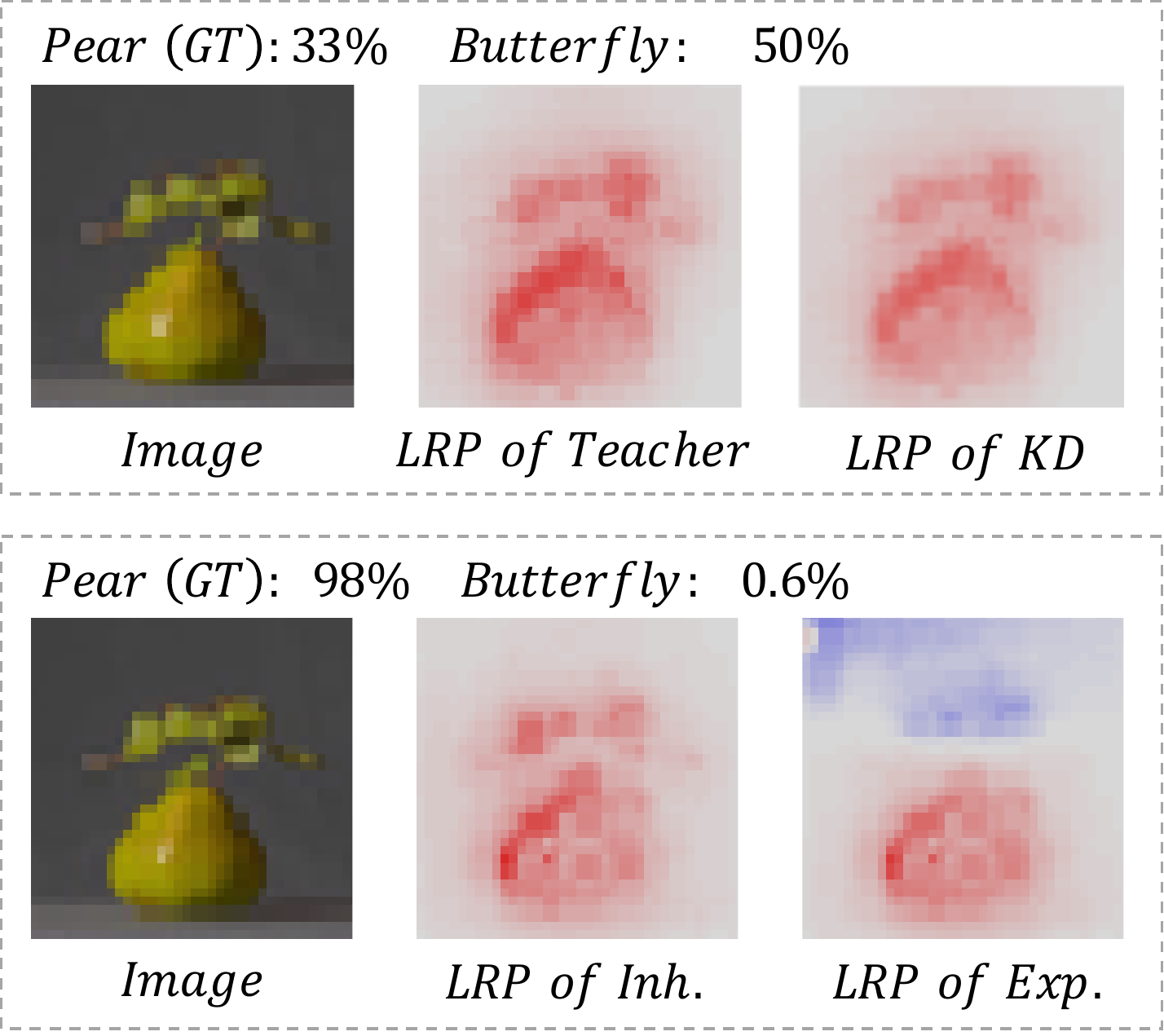}
        \label{fig:exploration_a}
    }
    \subfigure
    {
        \includegraphics[width=0.28\linewidth]{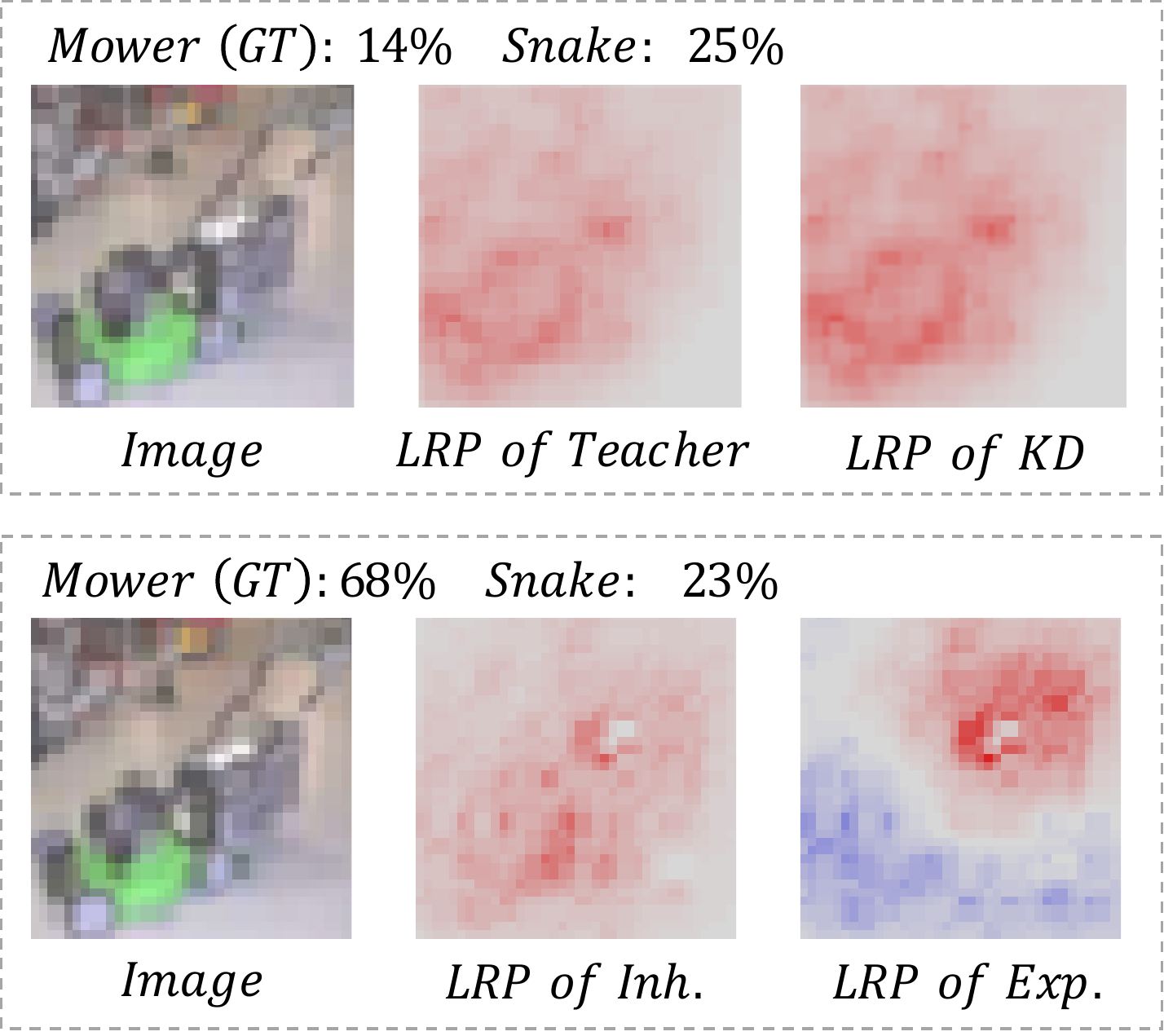}
        \label{fig:exploration_b}
    }
    \subfigure
    {
        \includegraphics[width=0.28\linewidth]{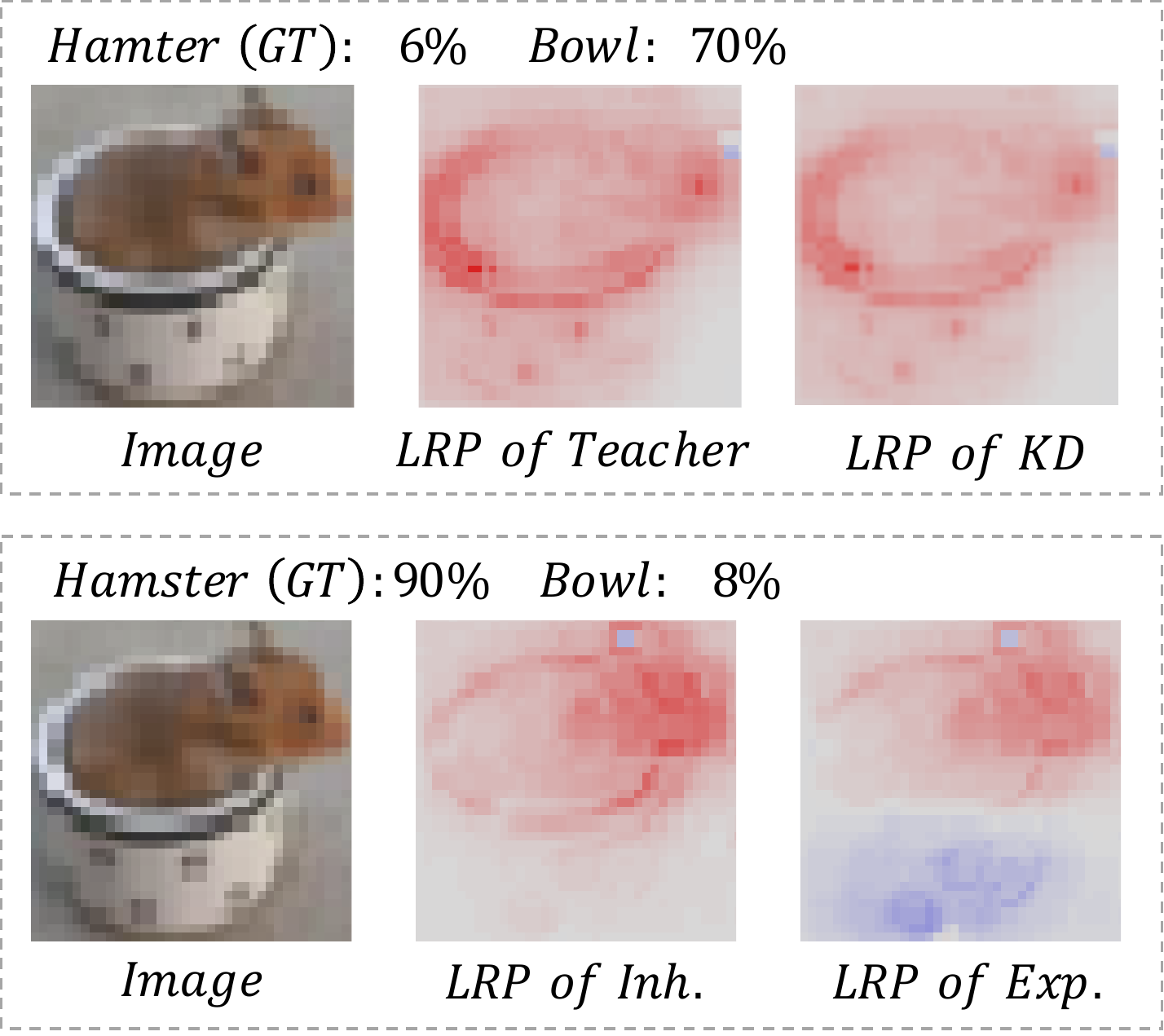}
        \label{fig:exploration_c}
    }
    \caption{
        Analysis on how exploration works. ``GT'' denotes the ground
        truth class of the image.
    }
    \label{fig:exploration}
    \vspace{-2mm}
\end{figure*}

\begin{figure*}[ht]
    \centering
    \vspace{-4mm}
    \subfigure[CKA]
    {
        \includegraphics[width=0.3\linewidth]{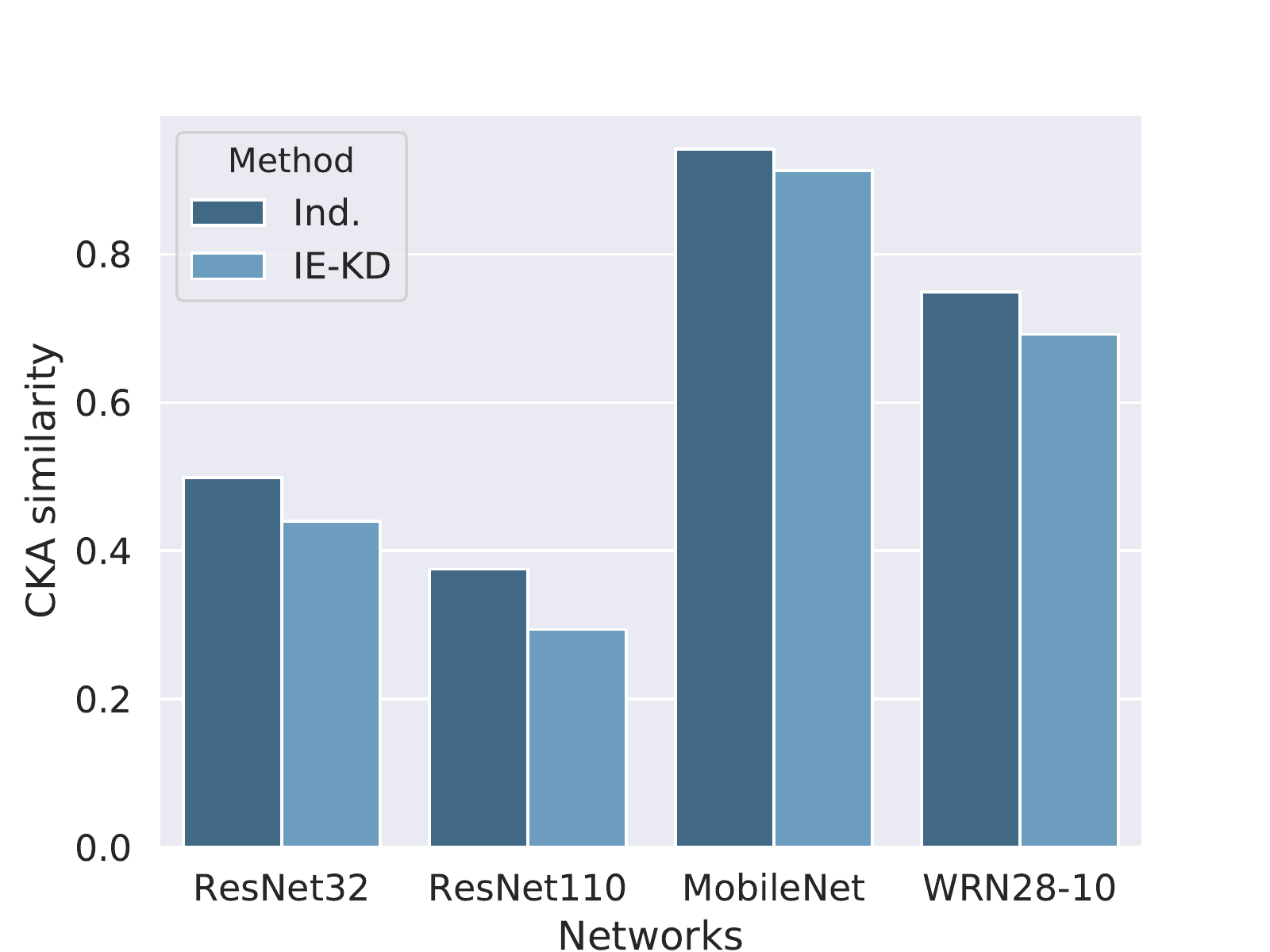}
        \label{fig:cka}
    }
    \subfigure[Active neurons]
    {
        \includegraphics[width=0.3\linewidth]{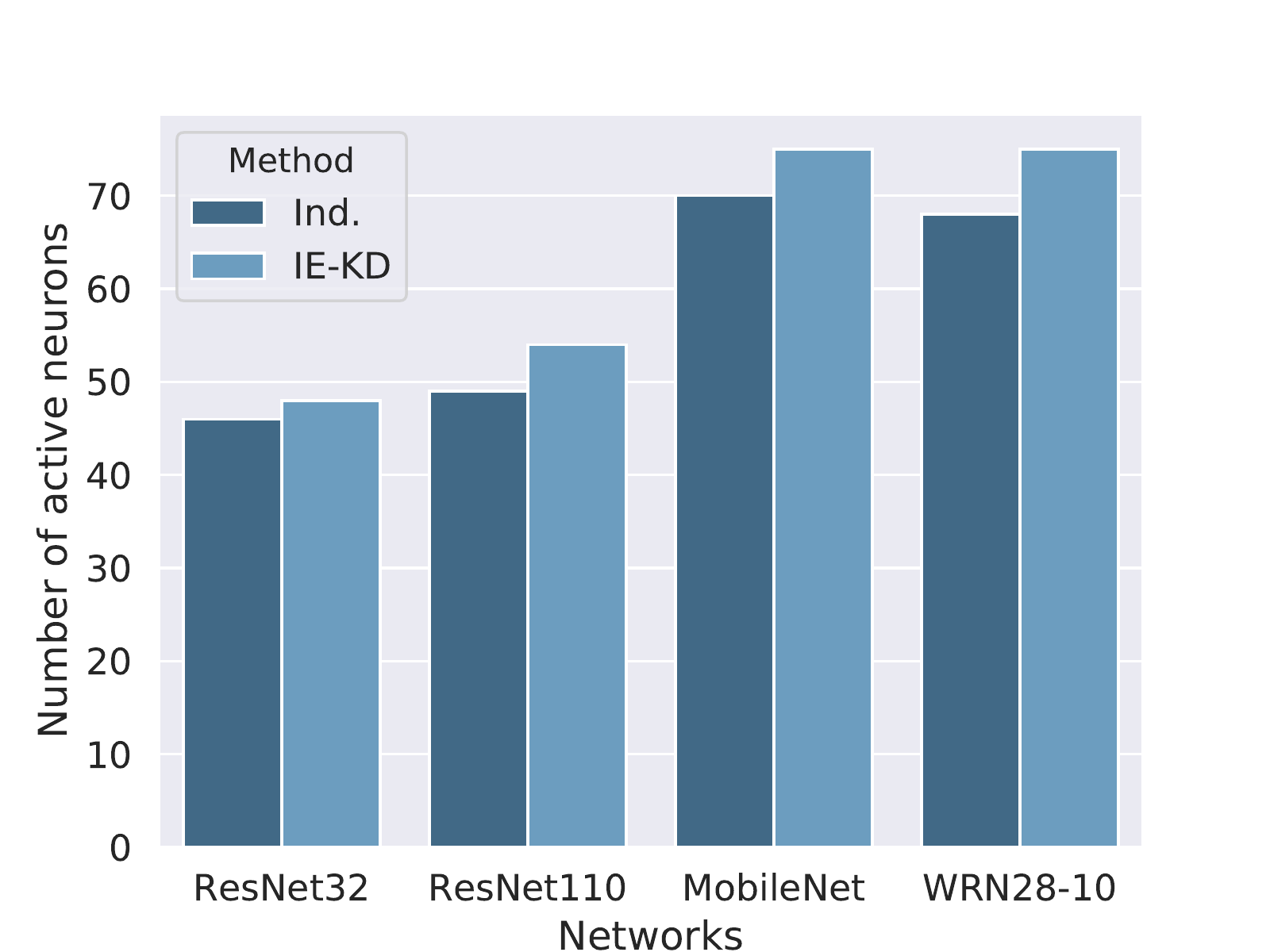}
        \label{fig:active_neurons}
    }
    \subfigure[Loss changes]
    {
        \includegraphics[width=0.32\linewidth]{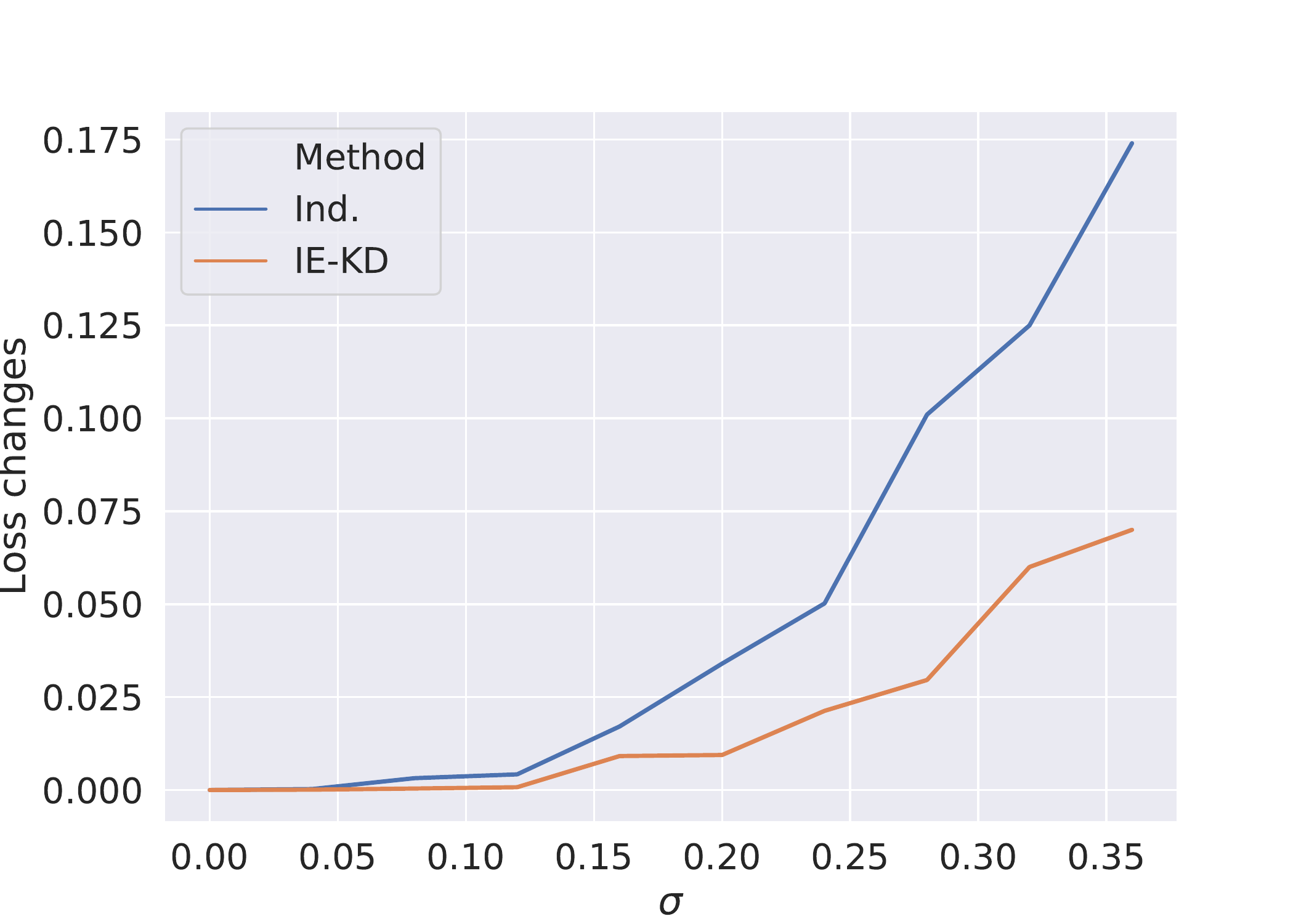}
        \label{fig:loss_changes}
    }
    \caption{
        Comparison of CKA similarity, numbers of active neurons, and loss changes when Gaussian noise is added.
    }
    \label{fig:analysis}
    \vspace{-3mm}
\end{figure*}

As discussed in Sec. \ref{sec:mutual_learning}, we can alternatively update the
student network by \LowRedistillation{} and the teacher network also by IE-KD
with the improved student network in a deep mutual learning manner
\cite{Zhang2017DeepML}, termed \emph{IE-DML}.
The learning rate strategy is the same as Sec. \ref{sec:settings}. The loss weights for reconstruction, inheritance and exploration loss are set as $0.8$, $50$ and $50$.
Table \ref{tab:cifar100-dml}
compares IE-DML with DML on CIFAR-100.
We experiment on two networks with
different depths (ResNet-32 / ResNet-110), different widths (ResNet32 / WRN-28-10),
different building blocks (ResNet-32 / MobileNet), and identical architectures
(ResNet32 / ResNet32). In all cases, IE-DML shows clearly better improvement than
DML and independent learning.  This implies that mutual learning of two
networks can also benefit significantly from our inheritance and exploration
framework. Our IE-KD is also a general upgrade of existing mutual learning
methods.
These results further confirm that IE-KD is a very general framework for
knowledge transfer between any type of networks and training strategies.

\subsection{Ablation Study}

\begin{table}[htpb]
    \vspace{-2mm}
    \centering
    \caption{
        Ablation study with different metrics and different loss weights on CIFAR-100. The teacher is
        ResNet-110 and the student is ResNet-32.
    }
    \subtable[metrics]{
    \begin{tabular}{cc}
        \toprule
        Metric & Error (\%) \\
        \midrule
        $L_1$ & \textbf{25.67} \\
        $L_2$ & 25.76 \\
        cos & 26.38 \\
        partial $L_2$ & 26.71 \\
        $L_1$+CKA & 26.60 \\
        \midrule
        Baseline & 31.01 \\
        \bottomrule
    \end{tabular}
    \label{tab:ablation_sim}}
    \subtable[loss weights]{
    \begin{tabular}{ccc}
        \toprule
        $\lambda_{inh}$ & $\lambda_{exp}$ & Error (\%) \\
        \midrule
        5 & 5 & 29.84 \\
        50 & 50 & \textbf{25.67} \\
        500 & 500 & 27.77 \\
        50 & 500 & 26.69 \\
        500 & 50 & 26.33 \\
        \midrule
        0 & 0 & 31.01 \\
        \bottomrule
    \end{tabular}
    \label{tab:ablation_weight}}
    \vspace{-2mm}
\end{table}

In the introduction section, we describe the motivation for a student network
to mimic the compact representation of a teacher network, and learn more
different and complementary features to improve diversity and generalization.
To further analyze the necessity and contribution of inheritance and
exploration, We conduct ablation studies to analyze the effects of inheritance
and exploration parts.

\textbf{Inheritance vs. Exploration.}
\quad
In Table \ref{tab:ablation_prop}, we show the results of using different
proportions of feature channels for inheritance and exploration, tested on
different datasets and network architectures. The settings of  ``$0.0 / 1.0$''
and ``$1.0 / 0.0$'' correspond to using either exploration or inheritance only.
The result shows that both inheritance and exploration are important, and even
division achieves the optimal performance.

\textbf{Metric.}
\quad
In Table \ref{tab:ablation_sim}, we show the results of using different
similarity metrics for the inheritance and exploration loss. For simplicity, we
use ``$L_2$'' to represent $L_2$ distance as similarity metric and its negative
counterpart as dis-similarity metric. The results show that no matter what
metric is used our IE-KD consistently improves the performance of the
student network.

\textbf{Loss weights.}
\quad
Table \ref{tab:ablation_weight} shows the results of using different
weights for the inheritance and exploration loss.
The results show that $\lambda_{inh}=\lambda_{exp}=50$ achieves the optimal
performance.

\subsection{How Does IE-KD Work?}
\label{sec:reasoning}

\textbf{Inheritance.}
\quad
We use layer-wise relevance propagation (LRP) \cite{lrpOverview} to interpret
the network by visualizing which pixels contribute how much to the
classification \cite{lrpProof}.  Fig. \ref{fig:exploration} shows different
images and their LRP heat maps from independent learning, KD and
\LowRedistillation{}.  We find that LRP heat maps of the inheritance channels
in IE-KD resemble the heat maps of the teacher network, which indicates that
the inheritance part mimics the compact features of the teacher network well.

Furthermore, we use the number of active neurons \cite{activeSubspace} to
analyze the redundancy of internal representations of networks.  For an
intermediate representation, the number of active neurons is the number of
directions to which classification loss function $c(x)$ is sensitive.  The more
active neurons, the less redundancy the representation contains
\cite{activeSubspace}.  The numbers of active neurons in the teacher network
(ResNet-110) and inheritance channels of student network (ResNet-32) are $46$ and $45$, indicating
that the inheritance channels contain approximately the same number of active
neurons as the teacher network. Moreover, the total number of neurons in
inheritance channels is only half of that in the teacher network, which means
that the knowledge is represented more compactly by the inheritance component.

\textbf{Exploration.}
\quad
First, we demonstrate that the exploration part can help discover more
\textit{discriminative} input patterns via some concrete examples.
In Fig. \ref{fig:intro_LRP}, the ``cheetah'' is misclassified as a
``crocodile'' by the student model, as the model attributes most of its
prediction to the tail of the ``cheetah'' that resembles a ``crocodile''. The
exploration part of \LowRedistillation{} model discovers new relevant patterns
on ears and mouth that are quite discriminative between the ``cheetah'' and
``crocodile'', and helps predict it correctly.
Fig. \ref{fig:exploration_a} shows another example, where the independently
trained student is confused by the leaf part of a ``pear'' and misclassifies
this image as a ``butterfly''.
The exploration part finds negative relevance of the leaf part (indicated by
blue) that pushes the model to focus more on the pear and less on the
leaf.
Similar result is shown in Fig. \ref{fig:exploration_b} and Fig. \ref{fig:exploration_c}. More results are
provided in the supplementary materials.

Second, we measure the similarity between features from the inheritance
and exploration channels, when student is trained independently or via
\LowRedistilling{}.  Centered Kernel Alignment (CKA) is introduced in
\cite{cka} as a similarity index to measure the similarity between two
representations. A larger CKA denotes a higher similarity between two sets of
representations For a fair comparison, the independent model and
\LowRedistilling{} model are initialized with the same random seed.
As shown in Fig.  \ref{fig:cka}, the CKA values are smaller for
\LowRedistilling{} in the four networks on CIFAR-100. This indicates more
diverse features in the \LowRedistilling{} model.

Third, we calculate the number of active neurons for \LowRedistilling{} and
independent student networks.
Fig. \ref{fig:active_neurons} demonstrates that
\LowRedistillation{} student network has more active neurons than the
independent student network. As proved in \cite{activeSubspace}, this means
that the features are more efficient.

\textbf{Generalization.}
\quad
Similar to \cite{Zhang2017DeepML,DML1,DML2}, we analyze the sharpness of the
converged minima of the independent and \LowRedistilling{} models in Fig.
\ref{fig:loss_changes}. Usually, sharp minima leads to poorer generalization,
while flat minima has better generalization ability\cite{DML1,DML2}. The
experiments are conducted on CIFAR-100 using MobileNet. The converged training
loss of both models is approximately the same, $0.131$ for IE-KD model
and $0.126$ for independent model, which means the depths of the two minima are
close. As we increase the scale of Gaussian noise added to model parameters,
the training loss of the independent model increases faster than that of the
IE-KD model. This suggests that the IE-KD model has found a much
flatter minimum, and it also provides another explanation for the lower
generalization loss of the IE-KD model in Fig.
\ref{fig:intro_loss}.

\section{Conclusion}
\label{sec:conclusion}

We propose a novel framework for neural network distillation, using the technique of inheritance and exploration. Our IE-KD framework is generic and can be easily combined with existing distillation or mutual learning methods.
Through experiments, we examine the performance of the proposed method using various networks in various tasks, and prove that the proposed method substantially outperforms the state-of-the-arts of knowledge distillation. We believe it can shed light on more future works, such as designing different forms of losses, or applying it to other tasks, \ieno, reinforcement learning.




\section*{Acknowledgements}
	This work was supported in part by Alibaba Innovative Research (AIR) program, Major Scientific Research Project of Zhejiang Lab (No. 2019DB0ZX01), NSFC No. 61872329, the National Key R\&D Program of China under contract No. 2017YFB1002203 and the Fundamental Research Funds for the Central Universities under contract WK3490000005.

{\small
\bibliographystyle{ieee_fullname}
\bibliography{egbib}
}

\end{document}


\title{Supplementary Material of ``Revisiting Knowledge Distillation: An
Inheritance and Exploration Framework''}

\author{
Zhen Huang\textsuperscript{1, 2}\thanks{This work was done when the author was visiting Alibaba as a research intern.},
Xu Shen\textsuperscript{2},
Jun Xing\textsuperscript{3},
Tongliang Liu\textsuperscript{4},
Xinmei Tian\textsuperscript{1}\thanks{Corresponding author.},\\
Houqiang Li\textsuperscript{1},
Bing Deng\textsuperscript{2},
Jianqiang Huang\textsuperscript{2},
Xian-Sheng Hua\textsuperscript{2}\footnotemark[2]\\
\textsuperscript{1}University of Science and Technology of China, 
\textsuperscript{2}Alibaba Group\\
\textsuperscript{3}University of Southern California,
\textsuperscript{4}University of Sydney\\
{\tt\small hz13@mail.ustc.edu.cn, junxnui@gmail.com, tongliang.liu@sydney.edu.au, \{xinmei,lihq\}@ustc.edu.cn,}\\
{\tt\small \{shenxu.sx,dengbing.db,jianqiang.hjq,xiansheng.hxs\}@alibaba-inc.com}
}

\maketitle



\section{Comparison with More Distillation Methods}

\begin{table*}[!htpb]
\centering
\caption{
    Top-$1$ test error (\%) of student networks on CIFAR100 of a number of
    distillation methods (ours are IE-AT, IE-FT and IE-OD) for knowledge
    transfer between different teacher and student architectures.
    IE-AT, IE-FT and IE-OD) outperform other distillation methods.
}
\begin{tabular}{ccccc}
    \toprule
    Teacher & ResNet-56 & WRN-40-1  & WRN-46-4 & WRN-16-2 \\
    Student & ResNet-20 & ResNet-20 & VGG-13   & WRN-16-1 \\
    \midrule
    Teacher & 6.39 & 6.84 & 4.44 & 6.27 \\
    Student & 7.78 & 7.78 & 5.99 & 8.62 \\
    \midrule
    KD \cite{Hinton2015DistillingTK} & 7.37 & 7.46 & 5.59 & 8.22 \\
    AT \cite{Zagoruyko2016PayingMA} & 7.13 & 7.14 & 5.48 & 8.01 \\
    FT \cite{Kim2018ParaphrasingCN} & 6.85 & 6.85 & 4.84 & 7.64 \\
    OD \cite{OD} & 6.81 & 6.69 & 4.81 & 7.48 \\
    Tf-KD \cite{KD-TF} & 7.41 & 7.51 & 5.48 & 8.10 \\
    CRD \cite{CRD} & 6.73 & 6.77 & 4.71 & 7.49 \\
    FitNet \cite{Romero2015FitNetsHF} & 7.23 & 7.24 & 5.50 & 8.19 \\
    SP \cite{SP} & 7.05 & 7.33 & 5.11 & 7.98 \\
    CC \cite{CC} & 6.98 & 7.05 & 5.29 & 7.89 \\
    VID \cite{VID} & 6.87 & 6.99 & 5.09 & 7.71 \\
    RKD \cite{RKD} & 6.93 & 6.92 & 4.91 & 7.82 \\
    PKT \cite{PKT} & 6.79 & 7.00 & 4.62 & 7.80 \\
    AB \cite{AB} & 7.01 & 7.29 & 5.31 & 8.04 \\
    NST \cite{NST} & 6.87 & 6.90 & 5.06 & 7.79 \\ 
    \midrule
    IE-AT & 6.70 & 6.81 & 4.55 & 7.76 \\
    IE-FT & 6.37 & 6.47 & 4.47 & 7.28 \\
    IE-OD & \textbf{6.33} & \textbf{6.39} & \textbf{4.45} & \textbf{7.16} \\
    \bottomrule
\end{tabular}
    \label{tab:KD}
\end{table*}

Using the technique of inheritance and exploration, our IE-KD framework is
generic and can be easily combined with existing distillation or mutual
learning methods.
Due to the space limitation, we only report the latest approaches in the main manuscript.

Here we show comparison with more distillation methods,
including KD \cite{Hinton2015DistillingTK}, AT \cite{Zagoruyko2016PayingMA}, FT
\cite{Kim2018ParaphrasingCN}, OD \cite{OD}, Tf-KD \cite{KD-TF}, CRD \cite{CRD},
FitNet \cite{Romero2015FitNetsHF}, SP \cite{SP}, CC \cite{CC}, VID \cite{VID},
RKD \cite{RKD}, PKT \cite{PKT}, AB \cite{AB} and NST \cite{NST}.
All experiments are conducted on CIFAR-10.
As the Table $1$ of the main manuscript,
we also use the four conditions to test various situations.
As shown in Table \ref{tab:KD}, all three variants (IE-AT, IE-FT, IE-OD) of our IE-KD consistently
outperform all previous distillation methods, regardless of the type of
teacher/student networks. 
These results further confirm that our IE-KD is a very general and effective
upgrade of existing distillation framework.

\section{Results of Small Teacher to Large Student}
\begin{table*}[!htpb]
\centering
\caption{
    Top-$1$ test error (\%) of the large student networks on CIFAR10 of a number of
    distillation methods (ours are IE-AT, IE-FT and IE-OD) for knowledge
    transfer from \textbf{small teacher network to large student networks}.
    IE-AT, IE-FT and IE-OD outperform other distillation methods.
}
\begin{tabular}{ccccc}
    \toprule
    Teacher & ResNet-20 & ResNet-20 & VGG-13   & WRN-16-1 \\
    Student & ResNet-56 & WRN-40-1  & WRN-46-4 & WRN-16-2 \\
    \midrule
    Teacher & 7.78 & 7.78 & 5.99 & 8.62 \\
    Student & 6.39 & 6.84 & 4.44 & 6.27 \\
    \midrule
    KD \cite{Hinton2015DistillingTK} & 7.16 & 7.38 & 4.71 & 6.83 \\
    AT \cite{Zagoruyko2016PayingMA} & 6.90 & 7.14 & 4.78 & 6.51 \\
    FT \cite{Kim2018ParaphrasingCN} & 6.38 & 6.85 & 4.84 & 6.32 \\
    OD \cite{OD} & 6.45 & 6.69 & 4.51 & 6.48 \\
    Tf-KD \cite{KD-TF} & 6.40 & 6.79 & 4.48 & 6.11 \\
    CRD \cite{CRD} & 6.68 & 6.77 & 4.51 & 6.31 \\
    FitNet \cite{Romero2015FitNetsHF} & 6.57 & 7.23 & 4.91 & 6.54\\
    SP \cite{SP} & 6.66 & 6.92 & 4.86 & 6.33 \\
    CC \cite{CC} & 6.58 & 6.97 & 4.76 & 6.24 \\
    VID \cite{VID} & 5.91 & 6.99 & 4.86 & 5.89 \\
    RKD \cite{RKD} & 6.92 & 6.89 & 4.57 & 5.98 \\
    PKT \cite{PKT} & 6.73 & 6.76 & 4.71 & 5.97 \\
    AB \cite{AB} & 6.75 & 7.02 & 4.70 & 6.26 \\
    NST \cite{NST} & 6.01 & 7.08 & 4.62 & 6.00\\ 
    \midrule
    IE-AT & 6.01 & 6.32 & 4.40 & 5.78 \\
    IE-FT & 5.65 & 6.90 & 4.31 & 5.62 \\
    IE-OD & \textbf{5.44} & \textbf{5.68} & \textbf{4.24} & \textbf{5.35} \\
    \bottomrule
\end{tabular}
    \label{tab:DKD-cifar10}
\end{table*}

\begin{table*}
\centering
\caption{
    Detailed architecture of the auto-encoder.
}
\begin{tabular}{cccccc}
    \toprule
    Layer & Kernel & Stride & Pad & \#Filters & Output\\
    \midrule
    Input & - & - & - & - & $N \times C \times H \times W$ \\
    Conv1 & $3\times3$ & 1 & 1 & $C$ & $N \times C \times H \times W$ \\
    Conv2 & $3\times3$ & 1 & 1 & $C/2$ & $N \times C/2 \times H \times W$ \\
    Conv3 & $3\times3$ & 1 & 1 & $C/2$ & $N \times C/2 \times H \times W$ \\
    Deconv1 & $3\times3$ & 1 & 1 & $C/2$ & $N \times C/2 \times H \times W$ \\
    Deconv2 & $3\times3$ & 1 & 1 & $C$ & $N \times C \times H \times W$ \\
    Deconv3 & $3\times3$ & 1 & 1 & $C$ & $N \times C \times H \times W$ \\
    \bottomrule
\end{tabular}
\label{tab:ae}
\end{table*}

\begin{table}[!htpb]
\centering
\caption{
    Top-$1$ and Top-$5$ test error (\%) of the large student networks on
    ImageNet of a number of distillation methods (ours are IE-AT, IE-FT and
    IE-OD) for knowledge transfer from \textbf{small teacher networks to large
    student networks}.
    The teacher is ResNet-18 and the student is ResNet-34.
    IE-AT, IE-FT and IE-OD outperform other distillation methods.
}
\begin{tabular}{ccc}
    \toprule
    & Top-1 & Top-5 \\
    Teacher & 29.91 & 10.68 \\
    Student & 26.73 & 8.57  \\
    \midrule
    KD \cite{Hinton2015DistillingTK} & 27.76 & 9.03 \\
    AT \cite{Zagoruyko2016PayingMA} & 27.01 & 8.81 \\
    FT \cite{Kim2018ParaphrasingCN} & 26.89 & 8.69 \\
    OD \cite{OD} & 26.77 & 8.66 \\
    Tf-KD(S) \cite{KD-TF} & 26.80 & 8.61 \\
    CRD \cite{CRD} & 26.55 & 8.55 \\
    FitNet \cite{Romero2015FitNetsHF} & 27.17 & 8.99 \\
    SP \cite{SP} & 26.89 & 8.72 \\
    CC \cite{CC} & 26.91 & 8.94 \\
    VID \cite{VID} & 26.88 & 8.78 \\
    RKD \cite{RKD} & 26.83 & 8.79 \\
    PKT \cite{PKT} & 26.76 & 8.63 \\
    AB \cite{AB} & 26.91 & 8.87 \\
    NST \cite{NST} & 26.90 & 8.75 \\
    \midrule
    IE-AT & 26.27 & 8.47 \\
    IE-FT & 25.83 & 8.29 \\
    IE-OD & \textbf{25.54} & \textbf{8.06} \\
    \bottomrule
\end{tabular}
    \label{tab:DKD-imagenet}
\end{table}

In the introduction of the main manuscript, we have described that simply
mimicking outputs of the teacher network will narrow the search space for
the optimal parameters of the student network and lead to a poor solution
from a feature learning view. This phenomenon becomes more evident when
transferring knowledge from a small teacher network to a large student network.

Thus, to further verify the effectiveness of our IE-KD framework, we validate its
performance on knowledge transfer from a small teacher network to a large
student network.
Similar to the settings in Table $1$ and Table $2$ of the main manuscript, the
architectures of student and teacher network are switched.
For example, the performance of regular knowledge transfer (a large teacher to
a small student) from ResNet-56 to ResNet-20 on CIFAR-10 is summarized in
Table \ref{tab:KD}.
Here, we consider the opposite direction that transferring knowledge from a
small teacher to a large student, \ieno, from
ResNet-20 to ResNet56.

Results of $3$ variants of IE-KD (IE-AT, IE-FT and IE-OD) and other
distillation methods on CIFAR10 and ImageNet are presented in Table
\ref{tab:DKD-cifar10} and Table \ref{tab:DKD-imagenet} respectively. We have
two observations.
1) previous distillation methods fail to bring consistent improvement when
the student network is larger than the teacher.
2) all variants of IE-KD show better performance than baseline student, as well
as other distillation methods.

These results show that our IE-KD is a general and effective framework, and can
improve the performance of student networks, regardless of the type of
teacher/student network and the relative size of teacher and student.

\section{Implementation Details of the Auto-Encoder}

In all our experiments, we used a simple auto-encoder having three
$2d$ convolution layers and three $2d$ transposed convolution layers.
All these six layers use $3 \times 3$ kernels, stride of $1$, padding of $1$, and batch
normalization with leaky-ReLU with rate of 0.1 followed by each of the six
layers. This means that we do not reduce spatial dimensions (height and width).
Instead, at the second convolution, we only decrease or increase the number of
output feature maps according to the number of factor feature maps.
Similarly, the second transposed convolution layer is resized to match the
feature maps of the teacher network.
The detailed architecture is illustrated in Table \ref{tab:ae}.
The auto-encoder is trained for the maximum of 30 epochs starting with learning
rate of 0.1.

\section{More LRP Examples}
\begin{figure*}[!h]
    \centering
    \subfigure[]
    {
        \includegraphics[width=0.3\linewidth]{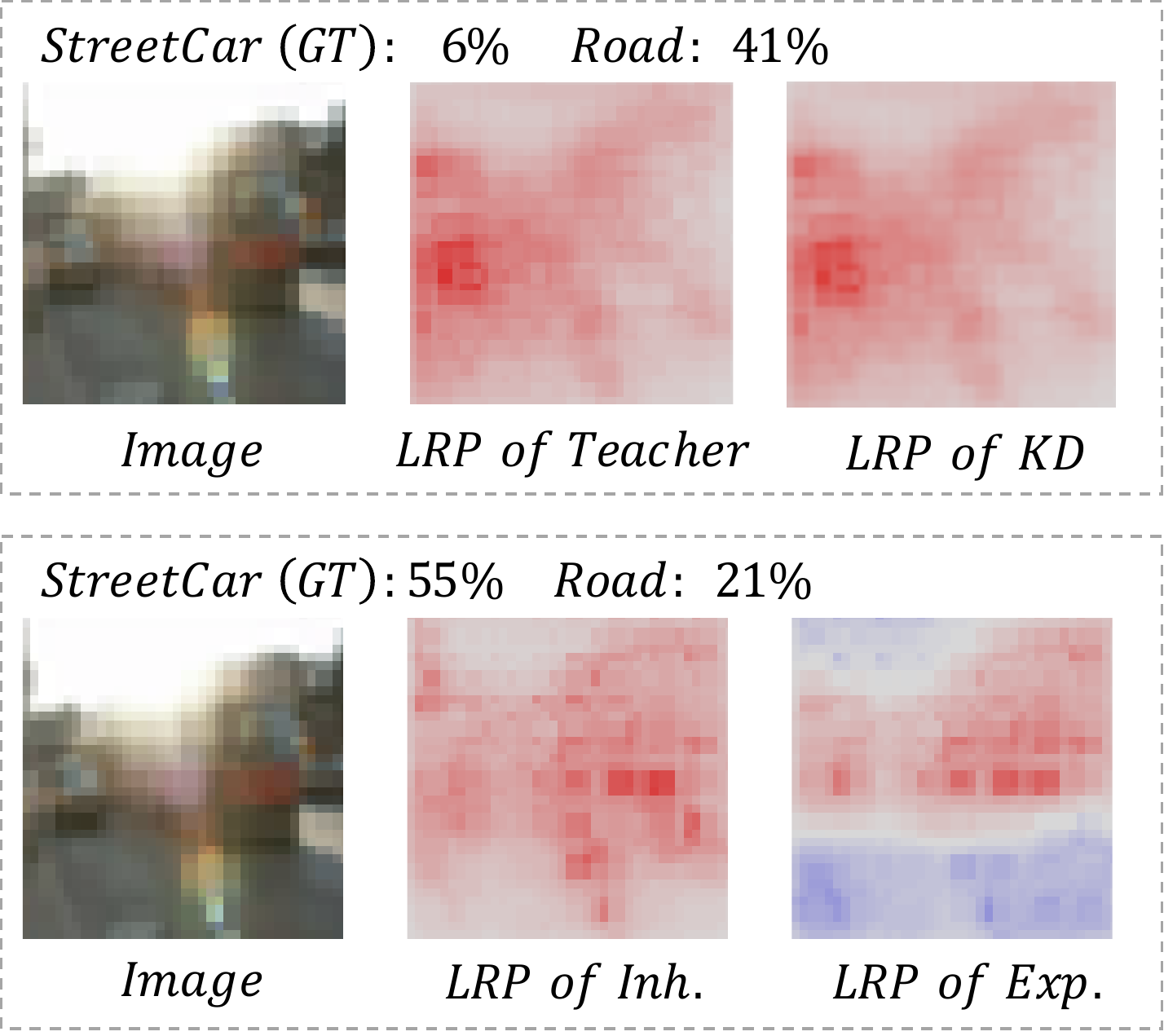}
    }
    \subfigure[]
    {
        \includegraphics[width=0.3\linewidth]{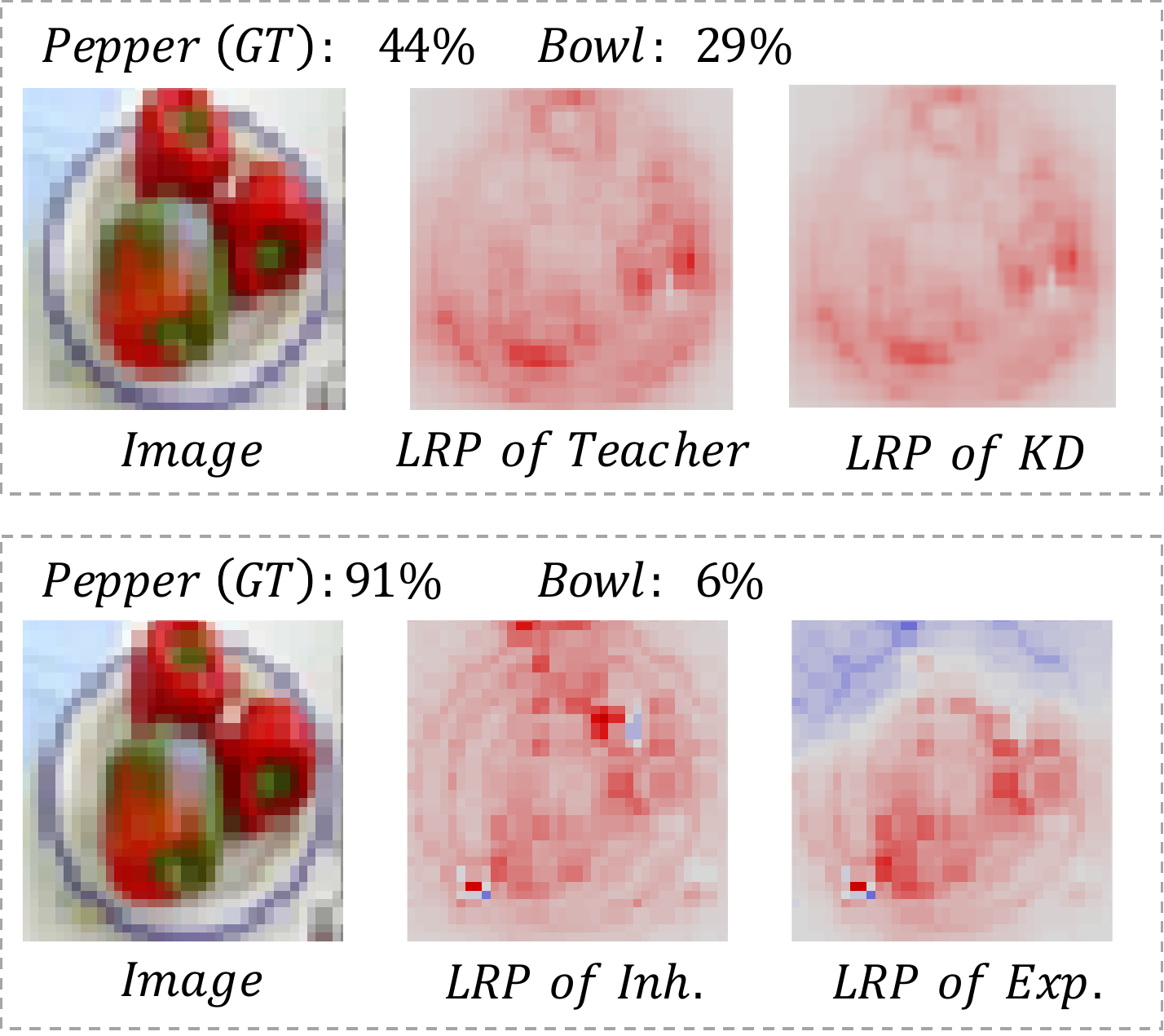}
    }
    \subfigure[]
    {
        \includegraphics[width=0.3\linewidth]{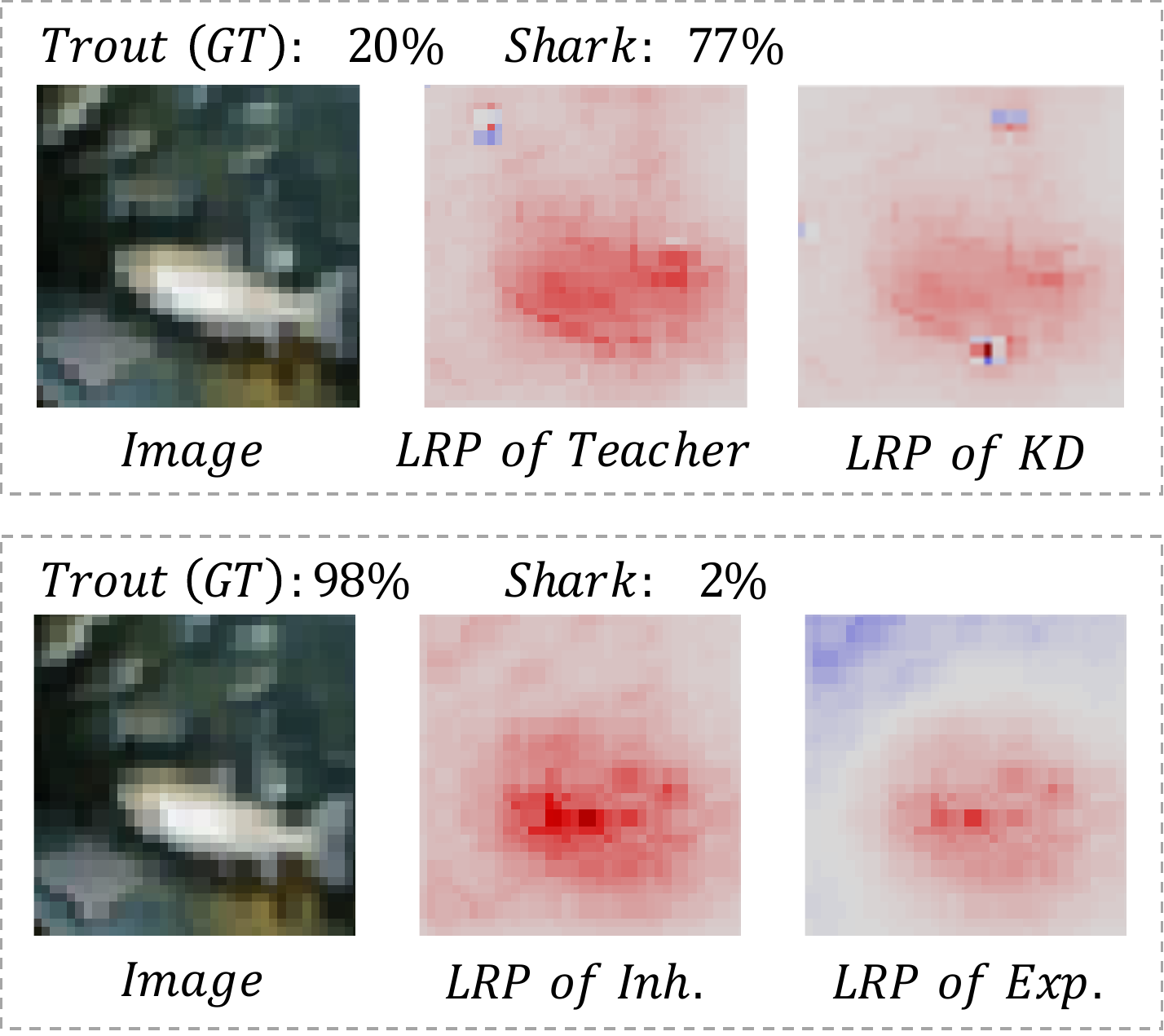}
    }
    \subfigure[]
    {
        \includegraphics[width=0.3\linewidth]{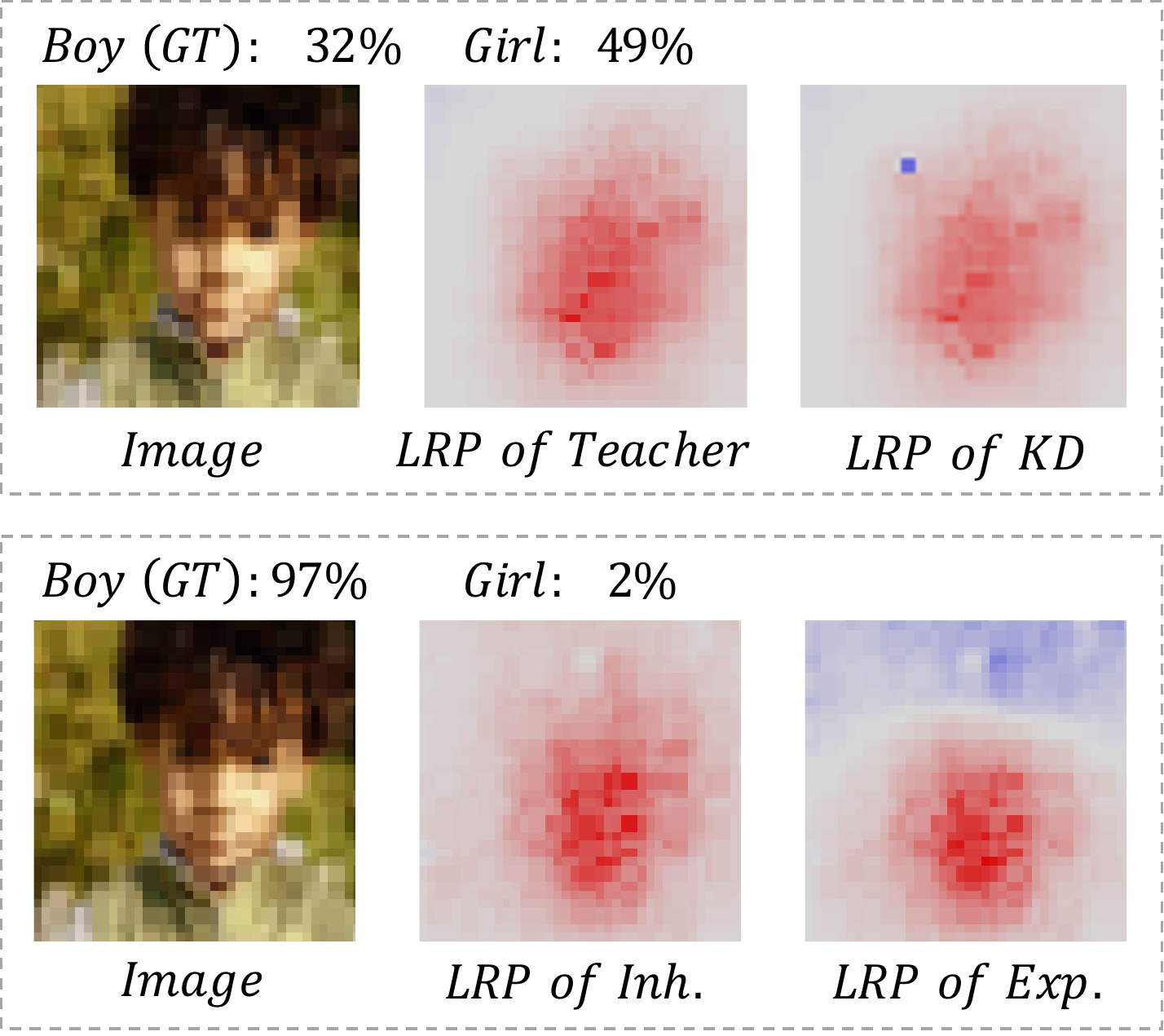}
    }
    \subfigure[]
    {
        \includegraphics[width=0.3\linewidth]{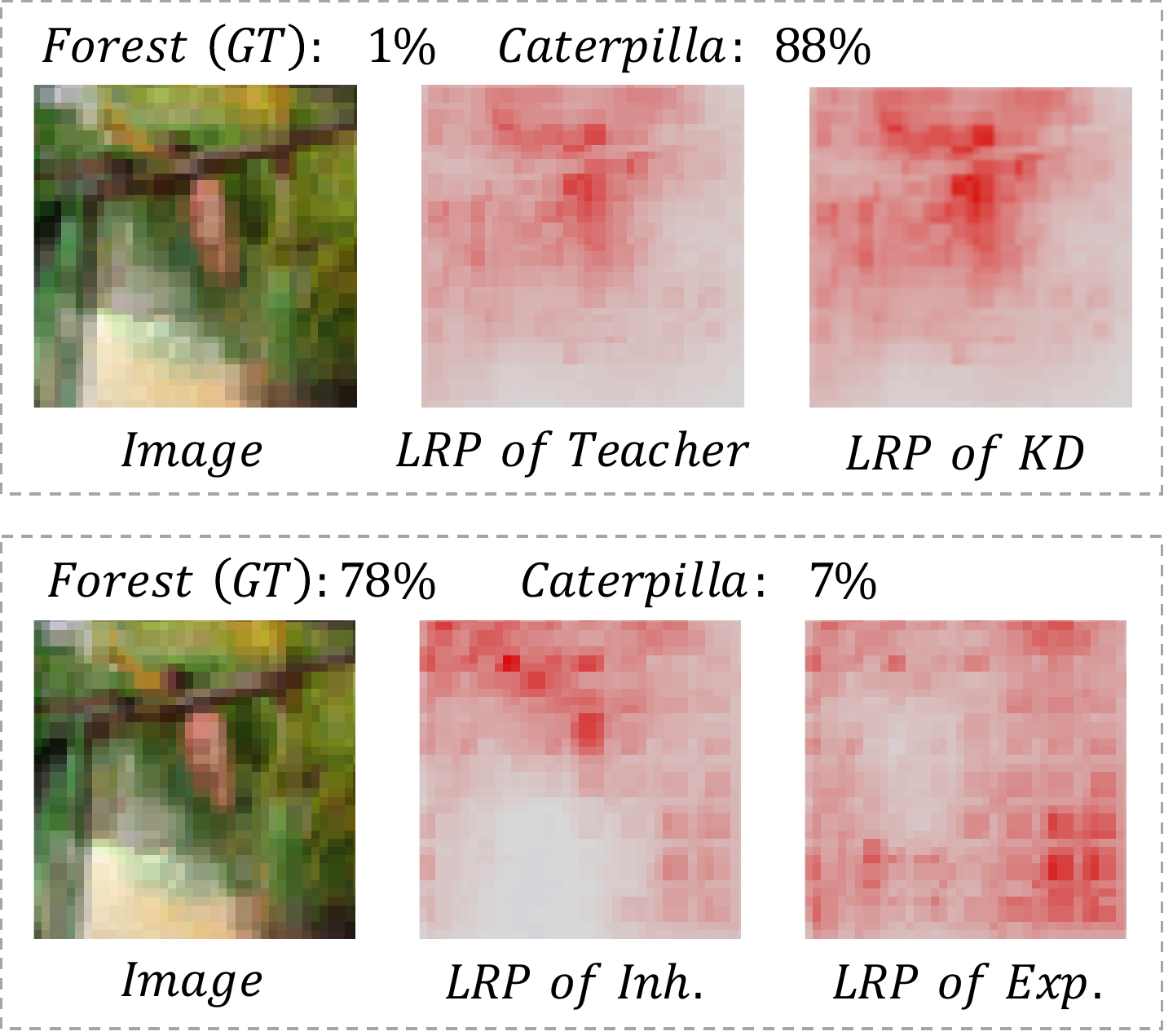}
    }
    \subfigure[]
    {
        \includegraphics[width=0.3\linewidth]{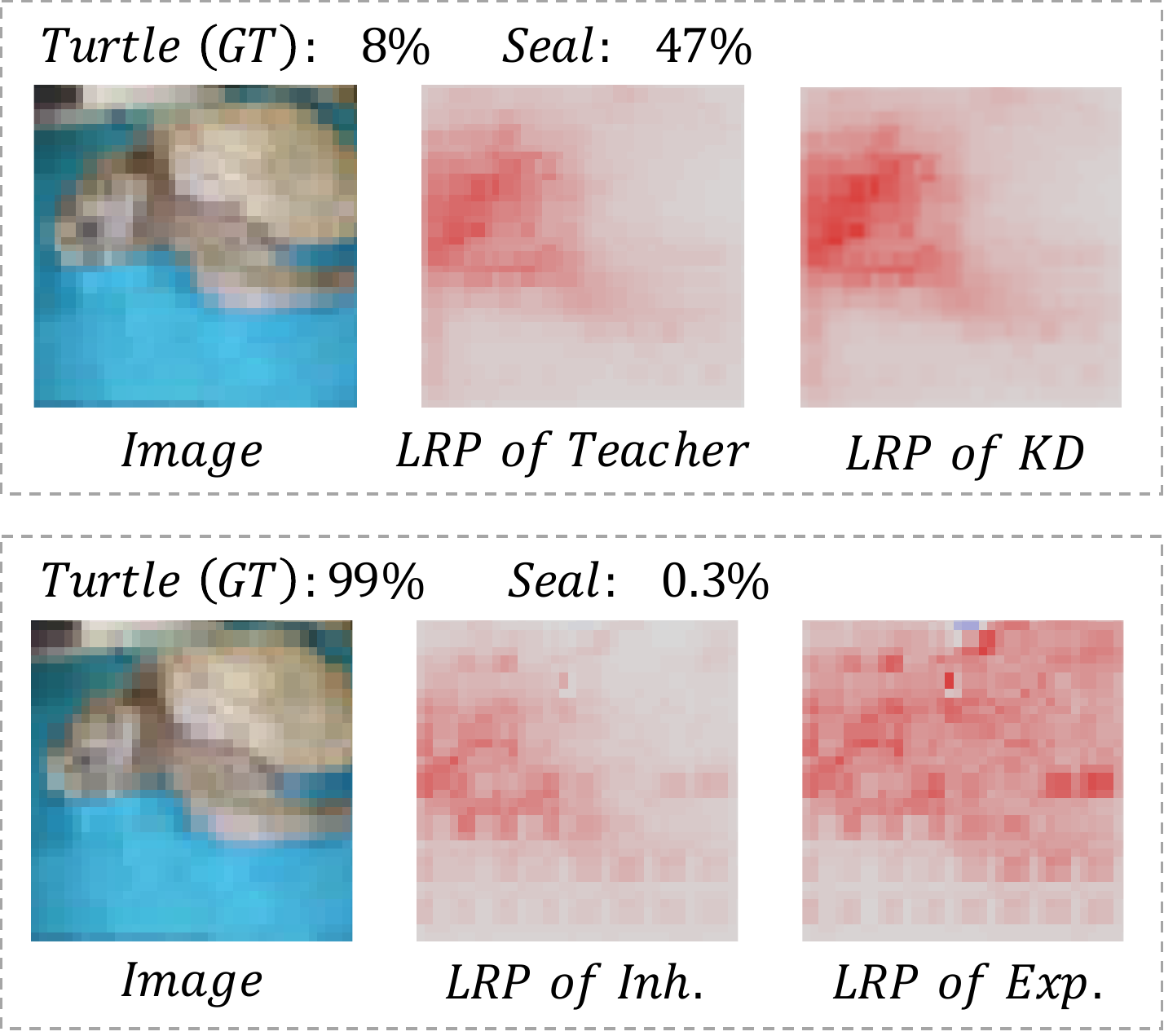}
        \label{fig:lrp_eg}
    }
    \subfigure[]
    {
        \includegraphics[width=0.3\linewidth]{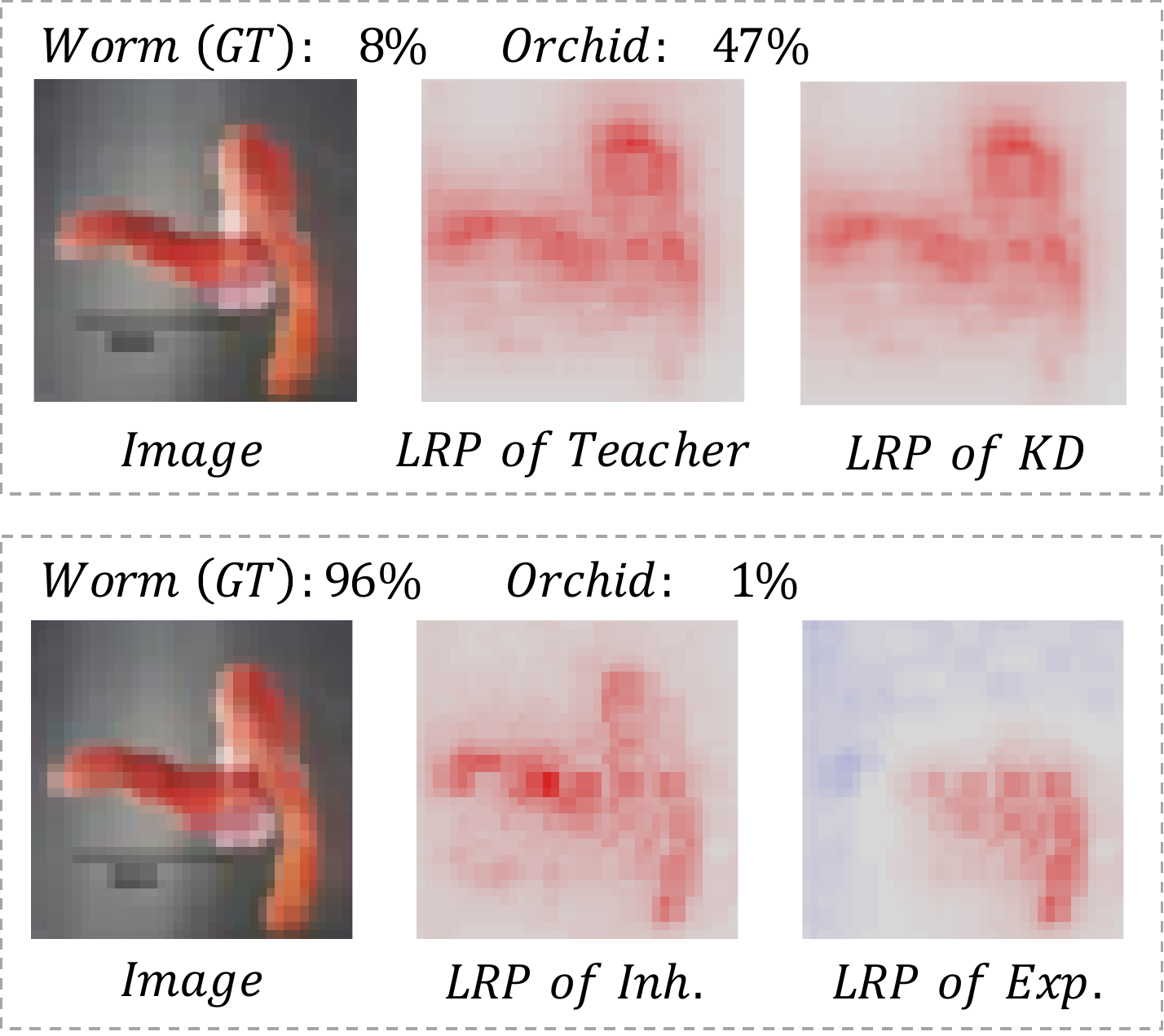}
    }
    \subfigure[]
    {
        \includegraphics[width=0.3\linewidth]{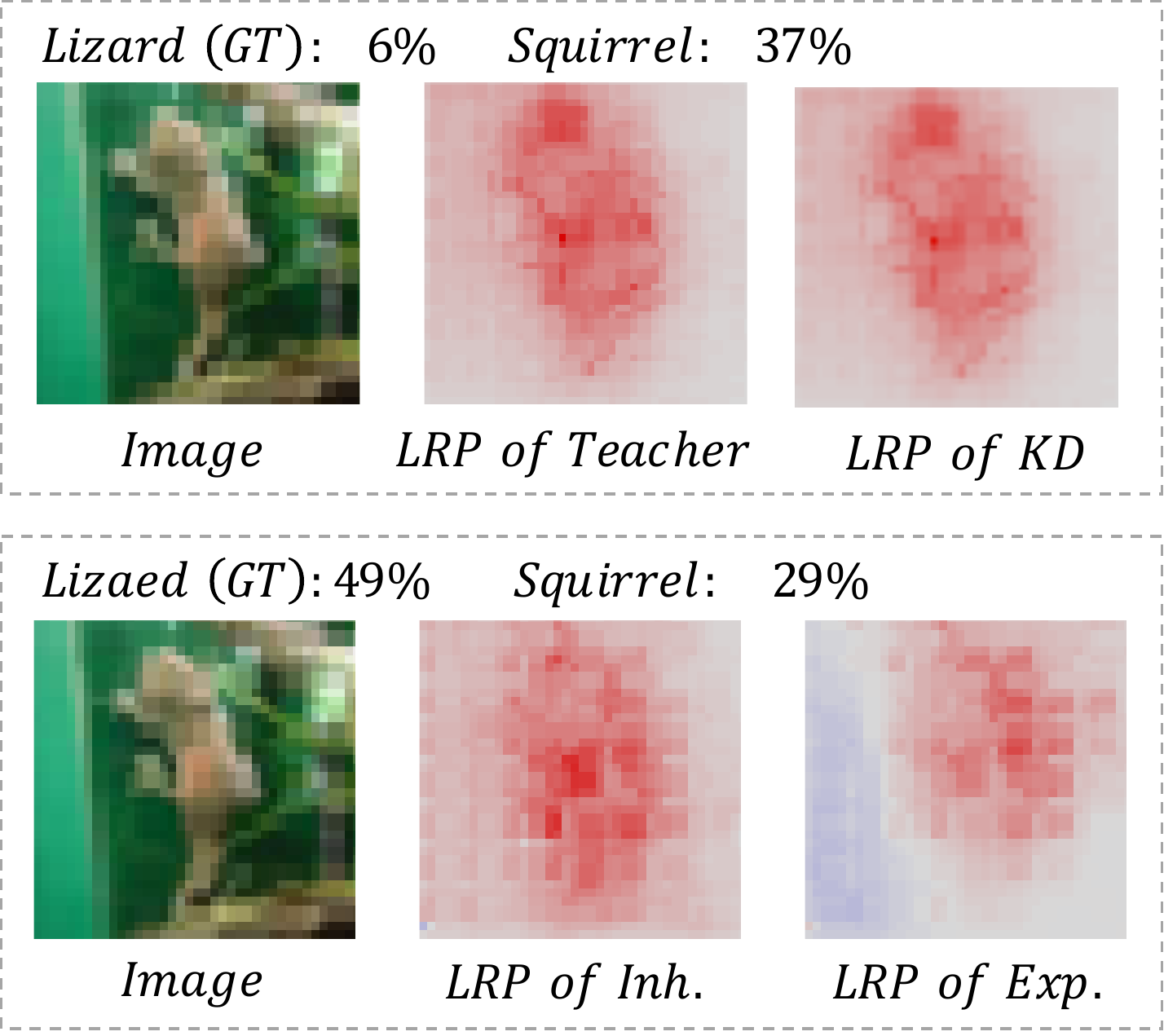}
    }
    \subfigure[]
    {
        \includegraphics[width=0.3\linewidth]{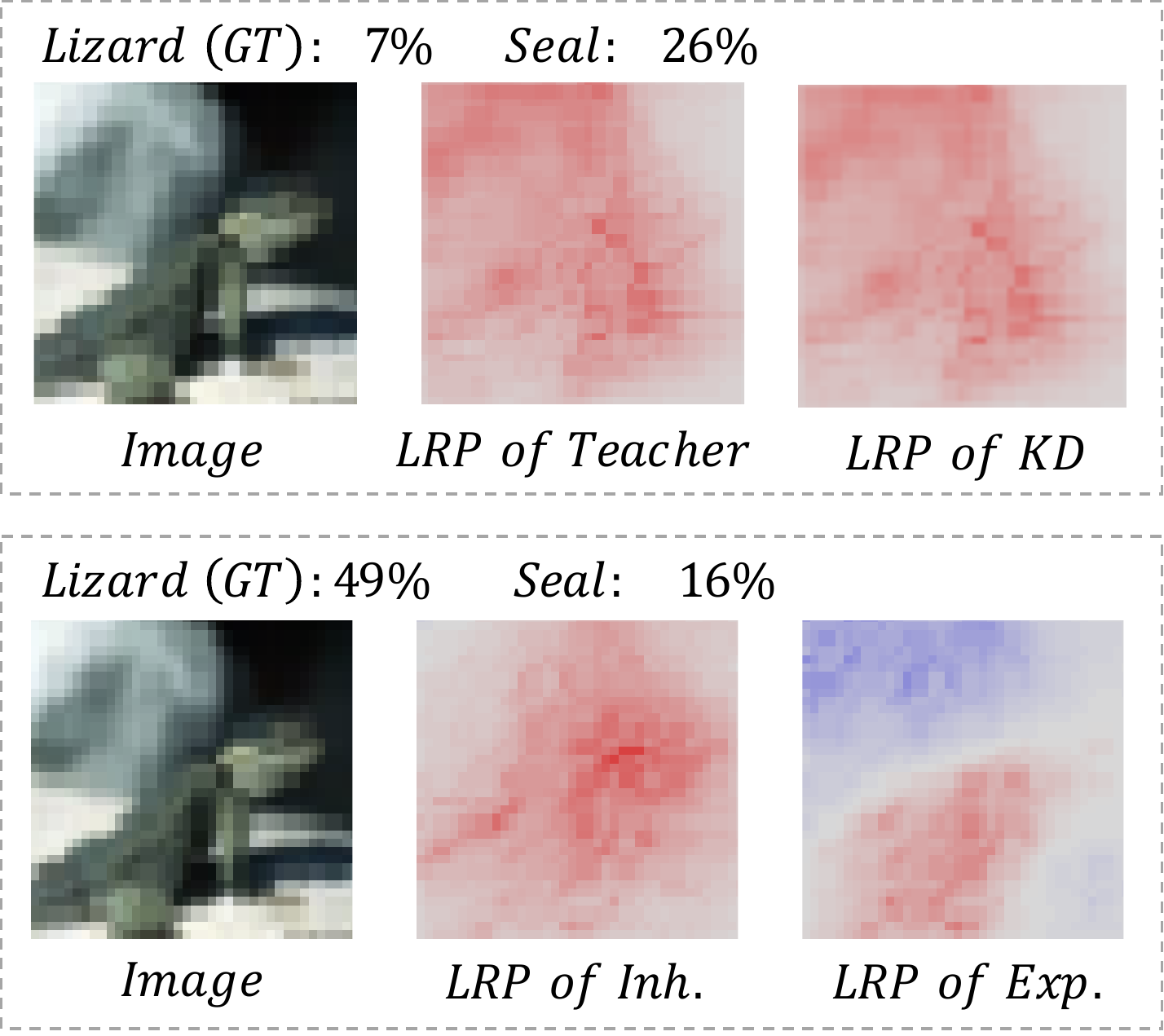}
    }
    \subfigure[]
    {
        \includegraphics[width=0.3\linewidth]{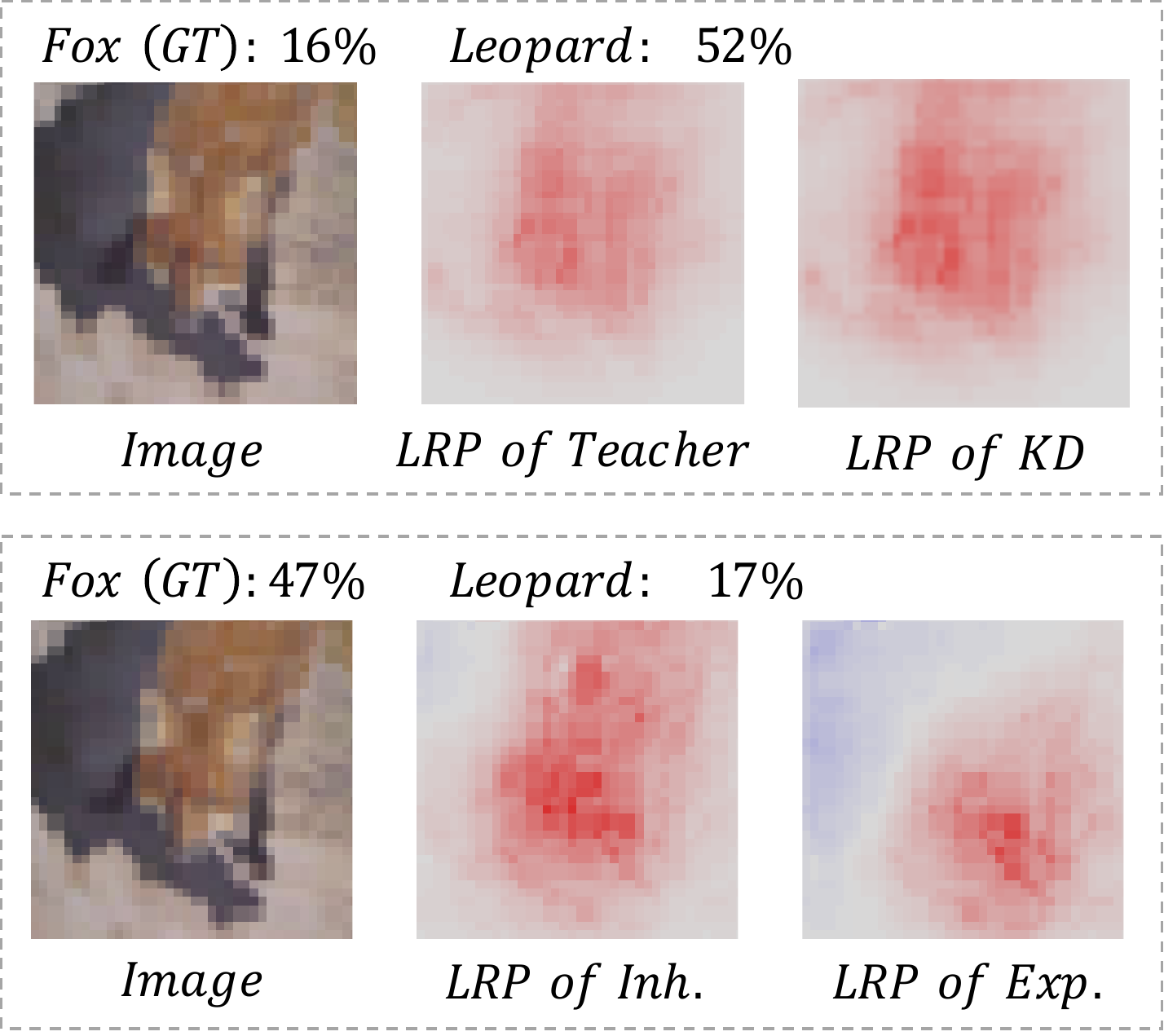}
    }
    \subfigure[]
    {
        \includegraphics[width=0.3\linewidth]{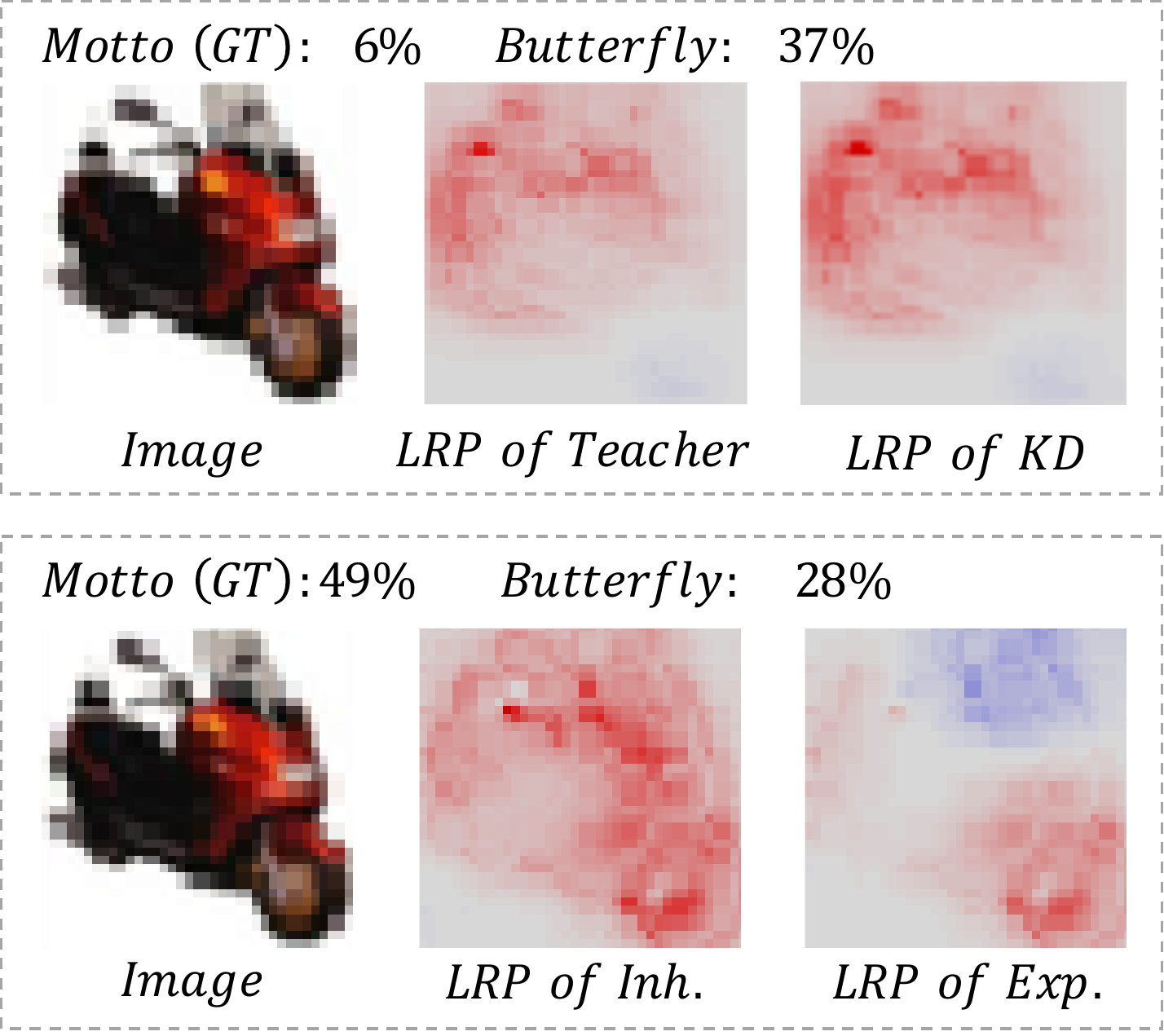}
    }
    \subfigure[]
    {
        \includegraphics[width=0.3\linewidth]{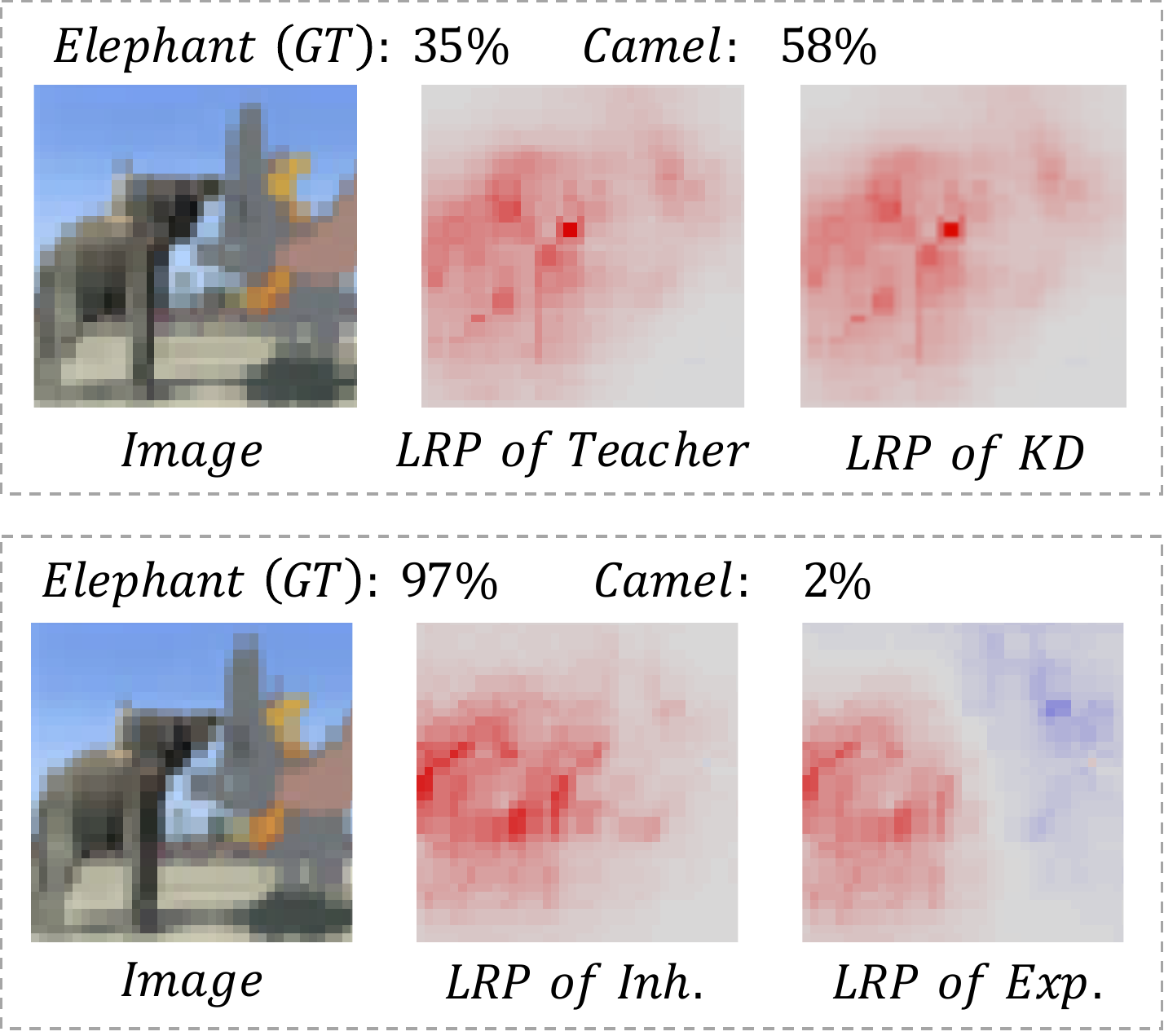}
    }
    \caption{
        More LRP cases for explaining how IE-KD works.
        ``GT'' denotes the ground truth class label of the image.
    }
    \label{fig:lrp}
\end{figure*}

Fig. \ref{fig:lrp} demonstrates more concrete LRP \cite{LrpOverview, LrpProof} examples. All of them
consistently illustrate that the exploration part helps to discover more
discriminative input patterns.
For example in Fig. \ref{fig:lrp_eg}, a ``turtle'' is misclassified as a “seal” by the
independent model, as the model attributes most of its relevance to the head of
the ``turtle'' that resembles a ``seal''.
The exploration part of IE-KD model discovers new relevant patterns on
the shell that are quite discriminative between the ``turtle'' and ``seal'', and
helps to make correct prediction.

\newpage
{\small
\bibliographystyle{ieee_fullname}
\bibliography{egbib}
}